\documentclass{article} 

\usepackage{amsmath,amsfonts,bm}

\def\eqref#1{equation~\ref{#1}}

\def\1{\bm{1}}

\DeclareMathAlphabet{\mathsfit}{\encodingdefault}{\sfdefault}{m}{sl}
\SetMathAlphabet{\mathsfit}{bold}{\encodingdefault}{\sfdefault}{bx}{n}

\usepackage{url}
\usepackage{pifont}
\usepackage{colortbl}
\usepackage{arydshln}

\usepackage{listings}
\usepackage{hyperref}
\usepackage{algorithm}
\usepackage{algorithmic}
\usepackage{multicol}

\usepackage{booktabs} 
\usepackage{multirow} 
\usepackage{cite} 
\usepackage{graphicx} 
\usepackage{xcolor} 
\usepackage{amsmath} 

\definecolor{lightyellow}{RGB}{255, 255, 204} 

\usepackage{mathtools}
\usepackage{multirow}
\usepackage{graphicx}
\usepackage{enumitem}

\usepackage[utf8]{inputenc} 
\usepackage[T1]{fontenc}    
\usepackage{hyperref}       
\usepackage{url}            
\usepackage{booktabs}       
\usepackage{amsfonts}       
\usepackage{nicefrac}       
\usepackage{microtype}      
\usepackage{amsmath}
\usepackage{dsfont}
\usepackage{multirow}

\usepackage{amssymb}

\usepackage{multicol}
\usepackage{subcaption}
\usepackage[font=small,labelfont=bf]{caption}

\usepackage[mathscr]{euscript}

\newtheorem{prop}{Proposition}[section]

\newcommand{\beq}{\begin{equation}}
\newcommand{\eeq}{\end{equation}}
\def\be {\begin{equation}}
\def\ee {\end{equation}}
\def\bs#1\es{\begin{split}#1\end{split}}
\def\ba#1\ea{\begin{align}#1\end{align}}
\def\baed#1\eaed{\begin{aligned}#1\end{aligned}}
\def\bged#1\eged{\begin{gathered}#1\end{gathered}}
\def\bea{\begin{eqnarray}}
\def\eea{\end{eqnarray}}
\usepackage{wrapfig}

\usepackage{hyperref}
\hypersetup{
colorlinks=true
,urlcolor=blue
,anchorcolor=blue
,citecolor=blue
,filecolor=blue
,linkcolor=blue
,menucolor=blue
,pagecolor=blue
,linktocpage=true
,pdfproducer=medialab
,pdfa=true
}

\usepackage{hyperref}
\hypersetup{
colorlinks=true
,urlcolor=blue
,anchorcolor=blue
,citecolor=blue
,filecolor=blue
,linkcolor=blue
,menucolor=blue
,pagecolor=blue
,linktocpage=true
,pdfproducer=medialab
,pdfa=true
}
\newtheorem{definition}{Definition}                             

\definecolor{mybrown}{RGB}{60, 80, 110}
\definecolor{mygray}{RGB}{135, 170, 190}
\definecolor{mybrown}{RGB}{40, 60, 90}
\definecolor{mygray}{RGB}{115, 150, 170}
\definecolor{mybrown}{RGB}{30, 45, 70}
\definecolor{mygray}{RGB}{95, 130, 150}

\definecolor{mybrown}{RGB}{25, 40, 80}
\definecolor{mygray}{RGB}{80, 120, 160}

\definecolor{mybrown}{RGB}{15, 30, 70}
\definecolor{mygray}{RGB}{60, 100, 140}
\definecolor{mygray}{RGB}{60, 80, 140}

\definecolor{dmybrown}{RGB}{10, 20, 50}
\definecolor{darkerMygray2}{RGB}{40, 50, 110}

\usepackage[accepted]{icml2024}

\title{SiT:   Symmetry-invariant Transformers for Generalisation in Reinforcement Learning}

\begin{document}

\twocolumn[
\icmltitle{SiT:   Symmetry-Invariant Transformers for Generalisation\\ in Reinforcement Learning}




\icmlsetsymbol{adv}{*}

\begin{icmlauthorlist}
\icmlauthor{Matthias Weissenbacher}{r}
\icmlauthor{Rishabh Agarwal}{adv,comp}
\icmlauthor{Yoshinobu Kawahara}{adv,r,u}
\end{icmlauthorlist}

\icmlaffiliation{r}{RIKEN Center for Advanced Intelligence Project, Tokyo, Japan}
\icmlaffiliation{comp}{Google DeepMind}
\icmlaffiliation{u}{Graduate School of Information Science and Technology, Osaka University, Japan}

\icmlcorrespondingauthor{Matthias Weissenbacher}{wbmatthias@gmail.com}

\icmlkeywords{Machine Learning, ICML}

\vskip 0.3in
]

\printAffiliationsAndNotice{\icmlEqualContribution} 

\begin{abstract}

An open challenge in reinforcement learning~(RL) is the effective deployment of a trained policy to new or slightly different situations as well as semantically-similar environments. 
We introduce {\bf S}ymmetry-{\bf I}nvariant {\bf T}ransformer (SiT), a scalable vision transformer~(ViT) that leverages both local and global data patterns in a self-supervised manner to  improve generalisation. Central to our approach is Graph Symmetric Attention,  which refines the traditional self-attention mechanism to preserve graph symmetries, resulting in invariant and equivariant latent representations. 
We showcase SiT's superior generalization over ViTs on MiniGrid and Procgen RL benchmarks,  and its sample efficiency on Atari 100k and CIFAR10.

\end{abstract}


\vspace{-0.1cm}
\section{Introduction}
Despite recent advances in reinforcement learning, out-of-distribution generalization remains an open challenge. A widely-used approach to improve generalisation in image-based RL is data augmentation \citep{NEURIPS2020_e615c82a,yarats2021image,hansen2021softda} but it can lead to over regularisation to specific augmentations. Moreover, data augmentation's inherent non-determinism can amplify the variance in regression targets, which can be detrimental to learning~\citep{hansen2021stabilizing}. 
Complimentary to data augmentation, leveraging symmetries can improve generalization and lead to sample-efficient RL~\citep{tang2021the,van2020mdp, weissenbacher2022koopman}. 

Image-based RL may benefit from both local  
and global symmetries, which preserve a particular structure or property within a neighborhood of a pixel or image patches and throughout the entire image respectively.
Enforcing local symmetries through data augmentation is sample inefficient and computationally expensive. When an image is divided into local patches to capture these symmetries, the number of augmented samples we may need to represent all possible variations grows exponentially. Given the prevalence of symmetries in RL settings, it is advantageous for neural networks to possess the capability to develop a understanding of these local and global symmetries in a self-supervised manner that is data-driven.



\begin{figure*}[t!]
\centering
 \includegraphics[width=\linewidth]{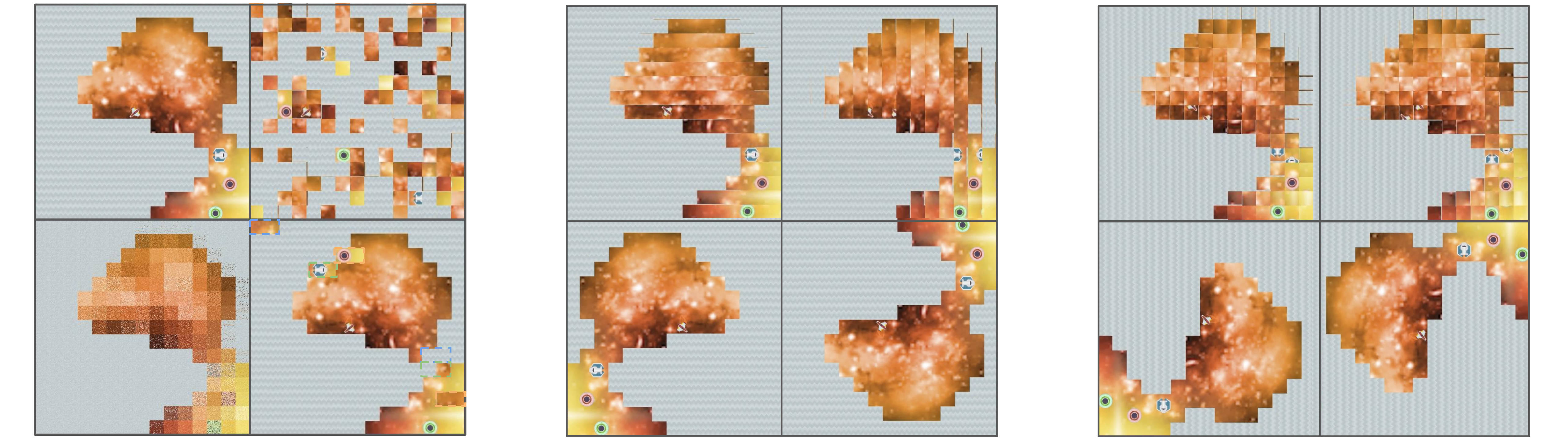}
    \begin{subfigure}{.32\linewidth}
    \vspace{0.1cm}
     \caption{Original (top-left) vs.\,patch and pixel-wise permutations.  Permutation result in non-solvable domains;   e.g. permuting 6 individual patches (bottom-right).}
      \end{subfigure}
      \hspace{0.35cm}
  \begin{subfigure}{.3\linewidth}
      \vspace{0.1cm}
    \caption{Horizontal and vertical  flips; patch-wise (top) - entire image (bottom). Approximate symmetry as it breaks left / right actions.}
    \end{subfigure}
     \hspace{0.4cm}
     \begin{subfigure}{.3\linewidth}
      \vspace{0.1cm}
    \caption{\footnotesize Left and right rotations by 90$^\circ$. Patch-wise (top): agent's local neighbourhood altered; Entire image: global exact symmetry (bottom).}
    \end{subfigure}
    \label{fig:walkersym- Sit}
    \vspace{-0.1cm}
    \caption{Local (patch-wise) and global transformations of observations of the CaveFlyer environment, Procgen suite \citep{cobbe20a}.  Permutation invariant agents \cite{tang2021the} can't discern key features (a) in contrast to agents with local and/or global flip and rotation  invariance   (b) and (c).
    }
    
     \vspace{-0.2cm}
    \label{fig:overview}
\end{figure*}

However, leveraging symmetries in RL presents various challenges. 
In particular, an agent's action choices in general are not invariant under symmetries both globally and locally, see Figure~\ref{fig:overview}. Permutation invariance~\citep{tang2021the} in Figure~\ref{fig:overview}~(a)  admits  the shortcoming that it leads to dead-end situations in many settings, while local and global flip symmetries (b) inter-changes left /right and up / down actions. Moreover, in many scenarios, it's essential for a decision-making process to consider the local context within the broader global setting, e.g., in Figure~\ref{fig:overview} (c),  the global 90$^\circ$ rotation is an exact symmetry but local patch-wise rotations change the neighbourhood of the agent.  In contrast, even minimal permutations are fatal for learning, see Figure~\ref{fig:overview}~(a) bottom-right.   
This situation is common amongst many  games and real-world environments \citep{bellemare2012investigating, cobbe20a,silver2016mastering, bellemare2020autonomous, kitano1997robocup}. 

To address the aforementioned challenges, we 
present a self-attention based network architecture, which we call 
\textbf{S}ymmetry-\textbf{i}nvariant \textbf{T}ransformer (SiT). 
SiTs incorporates a flexible relational inductive bias~\citep{DBLP:journals/corr/abs-1806-01261} 
to recognize relational patterns or symmetries, enabling it to adapt effectively to unfamiliar or out-of-distribution data. 
In addition to invariance, SiTs account for dead-end situations by incorporating equivariance, which refers to the property of an action to transform equivalently as the states under symmetries in our SiT module and by introducing a rotation symmetry preserving but flip-symmetry breaking layer.  Additionally, we introduce  novel invariant as well as equivariant  {\bf G}raph {\bf S}ymmetric {\bf A}ttention (GSA). GSA is akin to self-attention of Vision Transformers~(ViTs)~\citep{vit},
by adapting permutation-invariant self-attention \citep{pmlr-v97-lee19d} to maintain graph symmetries.  

SiTs capitalize on the interplay between local and global information. This is achieved by incorporating both local and global GSA modules. In particular, the local attention window stretches over several image patches, such that the local symmetries do not change the agent's local broader neighbourhood.
We demonstrate the efficacy of SiTs over ViTs on prevalent RL generalization benchmarks, namely MiniGrid and Procgen, and show sample-efficiency on the the Atari100k and CIFAR-10 vision benchmark.

In summary, our contribution is fourfold. First, we introduce a scalable invariant  / equivariant transformer architecture (SiT), i.e. accounting for symmetries down to the pixel level~(Section \ref{sec:sit}). Second, we perform empirical model evaluation on RL tasks; in contrast to conventional ViTs, SiTs require less hyper-parameter tuning, generalise better in RL tasks and are more sample-efficient. Specifically, SiTs lead to a $\mathbf{3\times}$ and $\mathbf{9\times}$ improvement in performance on commonly-used MiniGrid and Procgen environments. Third, SiTs incorporate a novel method  to account for the interplay of local and global symmetries, which is complementary to widely-used data augmentation in image-based RL. 
Fourth, GSA  is a novel approach to accomplish graph-symmetries in self-attention, not relying on positional embeddings \citep{NEURIPS2020_15231a7c, romero2021group}. We open sourced the SiT model-code on \href{https://github.com/matthias-weissenbacher/SiT}{ GitHub }.

\vspace{-0.1cm}
\section{Background}
\vspace{-0.1cm}
\label{sec:back}

\textbf{Reinforcement Learning.} A Markov Decision Process (MDP) is a mathematical framework for modeling decision-making problems in stochastic environments. MDPs are characterized by a tuple $(\mathcal{S}, \mathcal{A}, \mathcal{P}, \mathcal{R}, \gamma)$, where $\mathcal{S}$ is a finite set of states, $\mathcal{A}$ is a finite set of actions, $\mathcal{P}$ is the transition probability function, $R$ is the reward function, and $\gamma \in [0, 1)$ is the discount factor. 
In RL, one aims to learn optimal decision-making policies in MDPs. A policy, denoted as $\pi: \mathcal{S} \rightarrow \mathcal{A}$, is a mapping from states to actions. The optimal policy $\pi$  maximizes the expected cumulative discounted reward, given by the value function $V^\pi(s) = \mathbb{E}\left[\sum_{t=0}^{\infty} \gamma^t \mathcal{R}(s_t, a_t) \mid s_0=s, a_t \sim \pi(\,\,|s_t)\right]$, where the expectation is taken over the sequence of states and actions encountered by following the policy $\pi$. The optimal policy $\pi^{\ast}$ is the one that satisfies $V^{\pi^\ast}(s) \geq V^\pi(s)$ for all $s \in \mathcal{S}$ and any other policy $\pi$. 

\textbf{Invariance and Equivariance}. Invariance and equivariance are foundational concepts in understanding how functions respond to symmetries of their inputs. In RL, equivariance and invariance properties are imposed on the actor and value networks, e.g. in \cite{wang2023the} on top of SAC  \citep{SAC}. Before defining these concepts, we introduce some notation. The function \(f\) maps elements from space \(\mathcal{S}\) to \(\mathcal{S'}\)  and \(g\) denotes an individual transformation in a symmetry group. The functions \( \rho(g) \) and \( \rho'(g) \) describe the action of g   on spaces \(\mathcal{S}\) and \(\mathcal{S'}\), respectively, e.g. \( \rho(g) \cdot s \) signifies applying a transformation \( \rho(g) \) on an element \( s \) of \(\mathcal{S}\).
    \newline
    {\it Invariance:} A function \(f\) is invariant with respect to a set of transformations (symmetry group) if the application of any transformation from this set to its input does not change the function's output. Mathematically, this is expressed as: 
$ f(\rho(g) \cdot s) = f(s)$
for every transformation \(g\).
 {\it Equivariance:} A function \(f\) is equivariant if, when a transformation is applied to its input, there is a corresponding and predictable transformation of its output. This relationship is captured by the equation: $ f(\rho(g) \cdot s) = \rho'(g) \cdot f(s)$
for every transformation \(g \) in the symmetry group.
\textbf{Attention mechanisms.}  Recently, conventional self-attention have been employed in the context of RL  agents \citep{tang2021the}. The permutation invariant self-attention layer \citep{pmlr-v97-lee19d} uses a fixed Q-matrix (queries).
The original ViT architecture \citep{vit} naturally admits permutation invariance~(PI) due to use of token embeddings. PI is only broken by using the positional embedding \citep{romero2021group, NEURIPS2020_15231a7c}. 
The standard attention is  given by 
\vspace{-0.1cm}
\bea\label{eq:defatt2}
\text{Att}(K,V,Q) &=& \text{softmax}\Big(\; \tfrac{1}{\sqrt{d_f}} \, Q \, K ^T  \; \Big)\,  V \;\; ,
\eea
where \( K \), \( V \), and \( Q \) denote the keys, values, and queries respectively. They are derived from the input \( X \):
$ K = X W^k, \, V = X W^v, \, Q = X W^q$,
where \( W^q \), \( W^k \), and \( W^v \) are the corresponding weight matrices. 
The keys and values are constructed based on the input data, which is segmented into \( P \) patches. Consequently, the matrices \( K \), \( V \), and \( Q \) have dimensions \( \mathbb{R}^{P\times d_{f}} \), where \( d_{f} \) represents the feature dimension for each patch.

Graph neural networks and graph attention have been extensively explored in terms of their symmetries~\citep{vel2018graph, satorras_e_2021}. At a high-level, the graph attention mechanism (GAT) determines the relationships between nodes in a graph using attention.  The  attention  matrix  \eqref{eq:defatt2} is masked with the adjacency matrix $\mathcal{G}$ to ensure that the attention coefficients are only computed for nodes that are connected in the graph
\bea\label{eq:graphatt}
\text{GAT}\,(K,V,Q) &=&  \text{softmax} \Big( \tfrac{1}{\sqrt{d_f}}   \, Q \, K ^T \, \Big) \; \mathcal{G}\; V, \quad  
\eea
with $K,V,Q$  being the feature vectors of the  nodes, multiplied with weight matrices. A symmetrisation  of the score matrix may be added.\footnote{ Symmetrisation over the node / vertex indices given by $ \text{symmetric} (M) = M_{ij} + M_{ji}$ for $i,j = 1,\dots,P$ for a  square matrix $M \in \mathbb{R}^{P\times P}$.} to ensure that connections between nodes are bidirectional, meaning their importance is consistent regardless of direction.

\section{GSA: Symmetry-Invariant and Equivariant Attention}

\label{sec:graphsyms}

\begin{figure*}[t!]
 \includegraphics[width=\linewidth]{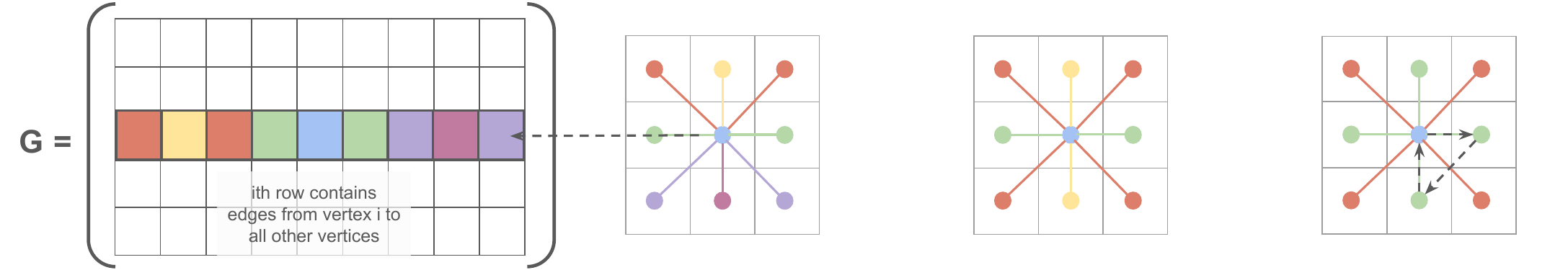}
   \vspace{0.1cm}
    \centering
    \begin{subfigure}{.42\linewidth}\vspace{0.2cm}
     \caption{Horizontally flip symmetry preserving Graph matrix.  Its i$^{th}$ column is given by the  horizontally flip symmetric graph centered around the i$^{th}$ patch vertex (here $i=5$).}
     \label{fig2a}
      \end{subfigure}
      \hspace{1.2 cm}
      \vspace{-0.1cm}
  \begin{subfigure}{.26\linewidth}
     \caption{Horizontal and vertical flip preserving graph. Additional rotation invariance requires green = yellow.  }
    \end{subfigure}
    \vspace{0.1cm}
     \hspace{0.10cm}
     \begin{subfigure}{.22\linewidth}
    \caption{Rotation preserving. Flip symmetry broken by sum over directed triangle sub-graphs.}
    \end{subfigure}
     \vspace{-0.2cm}
      \caption{Composition choices of the graph matrix $G \in  \mathbb{R}^{P \times P}$ for $P=9$ to  preserve different  symmetries. Same colours in $G$ represent  shared weights. In (c)  flips change the orientation of directed triangles i.e. clockwise to anti-clockwise while $90^\circ$-rotations preserve it.    }
    \label{fig:gsatt}
    \vspace{-0.1cm}
\end{figure*}

In this work, we propose a modification of the permutation invariant attention layer~\citep{pmlr-v97-lee19d}. This adaptation is specifically designed to respect the inherent symmetries of a square two-dimensional grid, which serves as our underlying graph structure. These symmetries include translations, rotations, and flips, as depicted in Figure ~\ref{fig:gsatt}. Our approach is an evolution of the rotary embedding method \citep{rotaryemb}. Our {\bf G}raph {\bf S}ymmetric {\bf A}ttention (GSA) layer is conceptually similar to a traditional graph-adjacency matrix.  Our graph topology matrix G is the analog of the adjacency matrix in \eqref{eq:graphatt} ; however,  its trainable weights are uniquely constrained to abide by certain symmetry conditions, which we discuss later.
 While our discussion  centers on the 2D grid, GSA may be adapted to 1D data where it ensures shift-symmetry (optionally flip-symmetry), see \ref{secapp:1D}.

For clarity, imagine a $9\times9$ pixel image. When segmented into $3\times3$ pixel patches, we get 9 distinct patches. In the \textbf{local} GSA setup, each graph vertex corresponds to an individual pixel, suggesting that in Figure~\ref{fig:gsatt}, the term "patches" is synonymous with pixels. In contrast, the \textbf{global} GSA interprets the image as a collection of $3\times3$ patches, where each patch's central point is symbolized by a graph vertex, aligning with the conventional ViT perspective. 
 Taking inspiration from self-attention in graphs, we propose {\bf G}raph {\bf S}ymmetric {\bf A}ttention (GSA):
\bea\label{eq:defatt3}
\text{GSA}(K,V,Q) = \text{softmax}  \Big( \tfrac{1}{\sqrt{d_f}} \, \Gamma(Q,K) \,  \,  \Big) \, {\color{brown} G_v} \; V  
\eea
with the attention score matrix given by
\bea  \nonumber  \Gamma(Q,K) = \text{symmetric} \Big(\;\;   \big(  \; {\color{brown}  G_q} \, Q \,\; [ \; {\color{brown}  G_k} \, K \;]^T\, \big)  \, \odot \, {\color{mygray}G}\, \;\;\Big) \;\;,
\eea
where  $ \odot $ is the point-wise Hadamard product. Here, $\Gamma$ is interpreted as the attention graph matrix of the underlying 2D pixel grid. Analogous to \eqref{eq:graphatt}, the grid symmetries are imposed by a graph topology matrix $G$ which breaks  permutation invariance of the standard self-attention~(\eqref{eq:defatt2}). Assuming that the image is split into $P$ patches, the graph matrices ${\color{brown} G_{k,v,q}} \,\in \, \mathbb{R}^{P \times P \times d_f}$ and ${\color{mygray}G} \,\in \, \mathbb{R}^{P \times P \times \text{\# heads} }$ are to be chosen for each feature/head from either of the different symmetry preserving graph matrices depicted  in Figure \ref{fig:gsatt}. The matrix and point-wise multiplication in \eqref{eq:defatt3} is applied per each feature and head dimension, respectively. 

In Figure~\ref{fig:gsatt}, we highlight variants of a 2D grid topology matrix $G$ preserving different symmetries, where \emph{same} colors represent a \emph{shared} weight. For example,  when using horizontal in \ref{fig2a}, vertices and edges of the same color are transformed into each other, creating a symmetry. Other transformations do not produce this effect.
Now, we define G formally. For more technical details, see Appendix~\ref{app:details}.
Assume that the distances are measured w.r.t. a specific vertex, e.g. the center one in Figure~\ref{fig:gsatt}, and edges can be viewed as vectors. Then, pick $G \in \mathbb{R}^{P \times P}$ such that a shared weight is present in $G$:
\begin{itemize}[topsep=0.0em,leftmargin=1em,itemsep=0.0em]
\item \ref{fig:gsatt}(a). When horizontal component of edges have the same magnitude (Horizontal flip-preserving)
\item \ref{fig:gsatt}(b). When the magnitude of the edges is same (Horizontal and vertical flip-preserving)
\item \ref{fig:gsatt}(c). When the distance between vertices is consistent. (Rotation preserving)
\end{itemize}

\paragraph{Flip symmetry breaking layer which preserves rotation symmetry.}  To preserve meaning of the direction of the agent \cite{zhao2023integrating}, it is imperative to break flip-symmetries, as such symmetries interchange the sense of left / right or up / down. To do so, we consider directed triangle sub-graphs. In Figure~\ref{fig:gsatt}(c), flips and rotation acting are symmetries as they map the graph $G$ to itself. However, the directed triangle changes orientations from clock-wise to counter-clockwise  for flips, while it remains the same for rotations. Applying this insight  to the attention score matrix  $\Gamma$  one  breaks flip symmetry  and preserves the rotation symmetry by summing over distinct directed triangle sub-graphs
 \bea
 \label{eq:rot}
&& \Gamma^{\text{rot}}(Q,K)_{ij} \;\; =\;\;  \Theta^{(i\to j \to k)}\, \Gamma(Q,K)_{ij} \\[0.2 cm] \nonumber
   &+& \,\Theta^{(j\to k \to i)} \,\Gamma(Q,K)_{jk} \, + \, \Theta^{(k\to i \to j)}\, \Gamma(Q,K)_{ki} \;\; ,
 \eea
where $\Theta^{(j\to k \to i)}$ are shared weights if  the triangle angels are the same; for vertices $j,k,i$.  The resulting new graph score matrix $\Gamma^{\text{rot}}$ distinguishes between flips of the input data, but is invariant under $90^\circ$ left and right rotations.


\begin{prop}[Symmetry Guarantee]
\label{thm:sym}
The GSA mechanism~(\eqref{eq:defatt3}) represents a symmetry-preserving module. It may be both invariant and/or equivariant w.r.t. symmetries of the input. The corresponding symmetry is dictated by the various graph selections. To achieve rotation invariance, the subsequent application of \eqref{eq:rot} is necessary. 
   For {\bf invariance}  the  token embedding i.e. the artificial (P-1)$^{th}$ patch is utilized at the output. Due to this mechanism, self-attention~(\eqref{eq:defatt2}) is permutation invariant.
{\bf Equivariance} is achieved for the P-dimensional patch information of the output, i.e.\, not related to the token embedding.

\end{prop}

        


\textbf{Explicit \& Adaptive Symmetry Breaking}.
    The  graph matrices  ${\color{mygray}G}$  and ${\color{brown} G_{k,v,q}}$ at weight initialisation  explicitly break the symmetry of self-attention from  permutation invariance~(PI) or equivariance~(PE) to the respective choice, see Figure~\ref{fig:gsatt}. However, PI or PE may be approximately recovered by GSA during training in a self-supervised manner, as it corresponds simply to the identity matrix ${\color{mygray}G}, \, {\color{brown} G_{k,v,q}},  = \bf{1}$. E.g. in  Figure \ref{fig:gsatt}(b), rotation invariance is obtained if the yellow weights approximate the green ones. 

As far as we know, this symmetry-preserving GSA has not appeared previously. Prior work~\citep{NEURIPS2020_15231a7c,romero2021group} discuss only modification of the positional embedding. The latter, is entirely omitted by our approach. Symmetries  in ~\citep{NEURIPS2020_15231a7c,romero2021group} are imposed by  addition of the positional embedding to the input, however  each subsequent attention layer is still PI invariant w.r.t.~to its respective inputs. In contrast, in our approach  each GSA layer is individually able to reduce PI invariance to graph symmetries and is thus able to infer spatial 2D information of latent features. 

\vspace{-0.1cm}
\section{Symmetry-Invariant and Equivariant Transformers }
\vspace{-0.05cm}
\label{sec:sit}

 \begin{figure*}[t!]
  \vspace{-0.2cm}
  \includegraphics[width=0.85\textwidth]{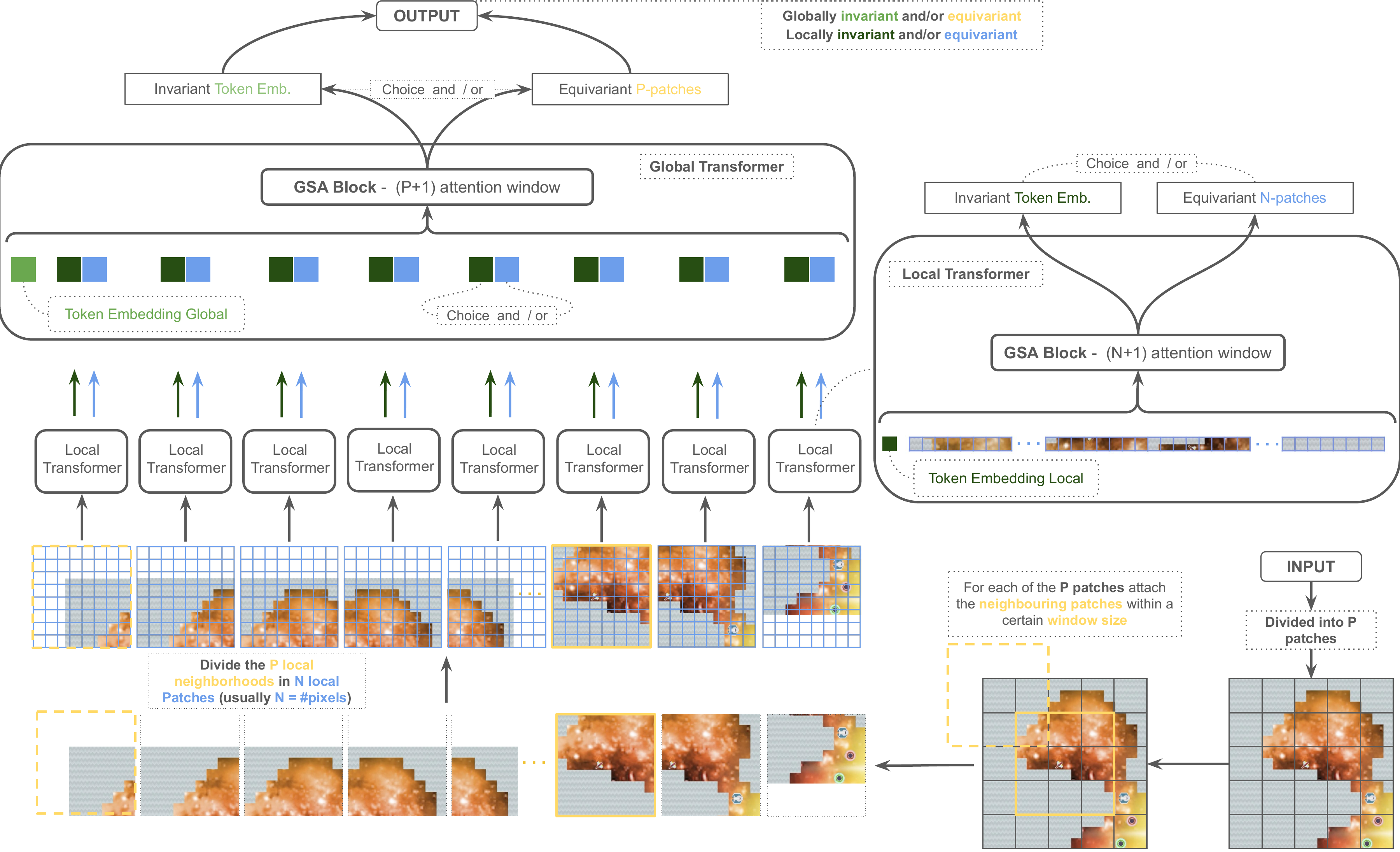}
    \centering
     \vspace{-0.2cm}
      \caption{SiT model architecture wit local and global GSA modules. }
    \label{fig:SitArchitecture_main}
    \vspace{-0.4cm}
\end{figure*}

Symmetry invariant transformer~(\textbf{SiT}) is a vision transformer that employs the GSA mechanism~(\eqref{eq:defatt3}), and optionally \eqref{eq:rot}, both locally as well as globally. See Figure~\ref{fig:SitArchitecture_main} for a visualization. We refer to attention applied to entire image patches as ``global''. On the other hand, ``local'' attention is applied to a specific patch or its surrounding neighborhood.  

Invariance is obtained by the same mechanism as the permutation-invariance of self-attention, i.e. the token-embedding is added as the (P+1)$^{th}$ patch to the input. Since the token embedding does not change under transformations of the input data, the transformer model remains invariant if only the token embedding is considered at the output. In contrast, the representation along the patch dimensions changes under symmetry transformations of the input; however there is a specific way in which one can trace that property throughout the transformer model; we refer to the latter as the equivariant patch-representation. For a more formal argument, please see appendix \ref{app:proof}. 

Based on the above discussion, an invariant SiT forward propagates only the symmetry-invariant token embedding to subsequent layers. In contrast, equivariant SiT~(\textbf{SeT}) forward propagates the equivariant patch-representation both locally and globally.
Symmetry-invariant-equivariant Transformer (\textbf{SieTs}) is both local and global invariant and equivariant.   The global symmetry are a result   of the local attention and the global attention mechanism. For example, a global 90$^\circ$ rotation is can be thought of as the rotation of the position of the patches (global GSA symmetry) and additionally local rotation of every single patch on pixel level (local GSA symmetry), see Figure~\ref{fig:SitSym}. 


\textbf{SiT with Preservation of Directions}.
Since all flips alter the interpretation of left/right and up/down, only local and  global rotation-invariant SiTs maintain the agent's meaning of direction. Additionally,  Figure~\ref{fig:overview}(c) illustrates that the agent's surroundings shift with local patch rotations. To address this challenge, we enlarge the local attention window across multiple patches. This "softly" breaks local symmetries, hence, only the global rotation persists as an exact symmetry in our empirically tested invariant SiT version in section \ref{sec:emp_eval}, the one most fitting for RL tasks. Nonetheless, local symmetries may be restored during training in a self-supervised manner.

 \textbf{Graph Symmetric Dropout}: A conventional dropout function  -  likely required for large  SiT models -  breaks the inductive bias explicitly. We introduce graph symmetric dropout which preserves the symmetry of SiT. A symmetry preserving dropout for the GSA layer is obtained by setting specific shared weights in the graph matrices ${\color{mygray}G}, \, {\color{brown} G_{k,v,q}} $ to zero. This statement follows from proposition \ref{thm:sym}.

\vspace{-0.1cm}
\subsection{Scalability of SiTs}
\label{sec:scaleSit4}
\vspace{-0.1cm}
ViTs are known to require more working memory (RAM) of the GPU than CNNs, due to the softmax operation \citep{dao2022flashattention}.
The local attention mechanism of SiTs is applied to larger  effective batch-sizes as the actual batch-size of the input is compounded by the number of total patches of the global attention. Using a larger local attention window only increases this overflow.  In our current implementation, SiTs are 2x-5x times slower to execute than ViTs of comparable size. However, this limitation is due to our custom implementation of our neural-net layers (GSA, graph triangulation) and may be resolved by a future custom CUDA implementation as SiTs can outperform ViTs that contain much larger number of trainable parameters than SiTs.

Nonetheless, technical obstacles arise when scaling SiTs to larger image and batch-sizes in image-based RL environments such as Procgen. We address these by modifying  the SiT implementation. First, we  establish a connection between graph matrices and depth-wise convolutions with graph-weights as kernels. 
Secondly, to accommodate for an extended local attention window the graph matrix connects pixels over lager distances  while the actual attention-mechanism is focused on a smaller patch.

\section{Empirical evaluation}

\label{sec:emp_eval}

\subsection{Gridworld}

\begin{figure}[t]
    \centering
    \includegraphics[width=0.5\textwidth,height=0.24\textwidth, keepaspectratio=false ]{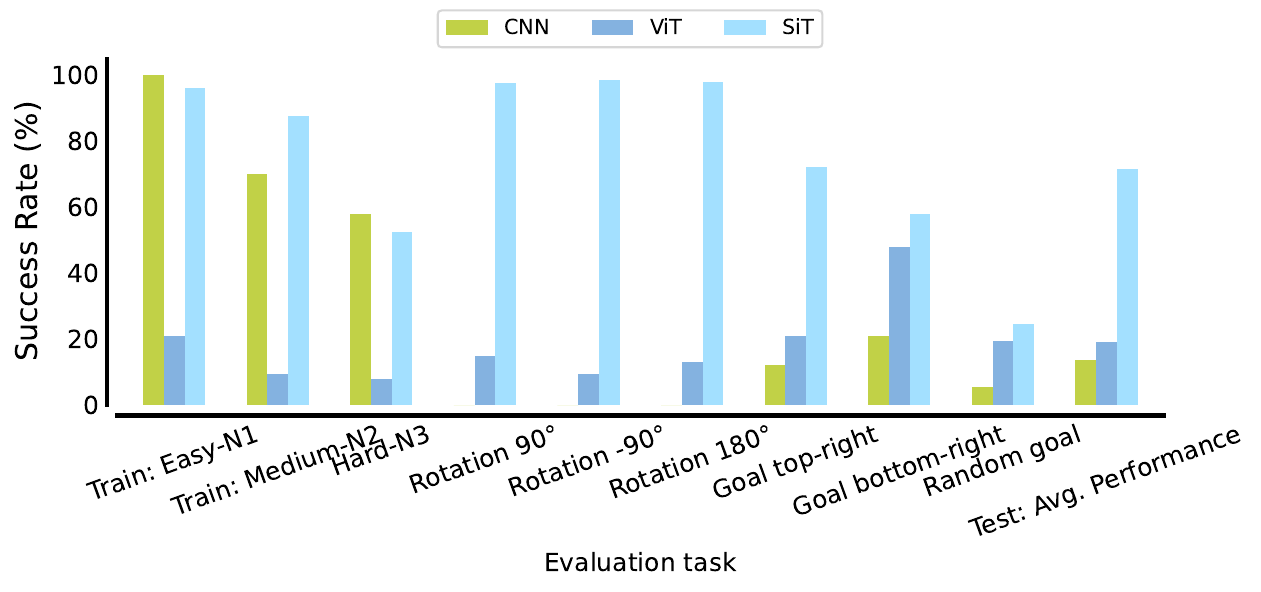}
    \vspace{-0.4cm}
    \caption{\textbf{Comparing SiTs with CNNs and ViTs}, in terms of training and generalization performance on LavaCrossing environments. SiTs substantially outperform both CNNs and ViTs.}
    \vspace{-0.1cm}
    \label{fig:mainres}
\end{figure}

\textbf{Environment Details}. The LavaCrossing environment is a standard component of MiniGrid, a Minimalistic Gridworld toolkit \citep{chevalier2018babyai}. The primary objective of the agent in this environment is to reach the goal position (green square) without falling into the lava river (orange squares). The game is procedurally generated with three levels of difficulty for each map size. Therefore, this environment is suitable for evaluating the combinatorial and out-of-distribution generalization of learned policies in RL. Moreover, we test the generalisation to rotated observations as well as goal changes, see (Figure~\ref{fig:lavacrossing}).
For further environment details see the appendix \ref{sec:hype}.

\begin{figure}[t]
    \centering
  \includegraphics[width=\linewidth]{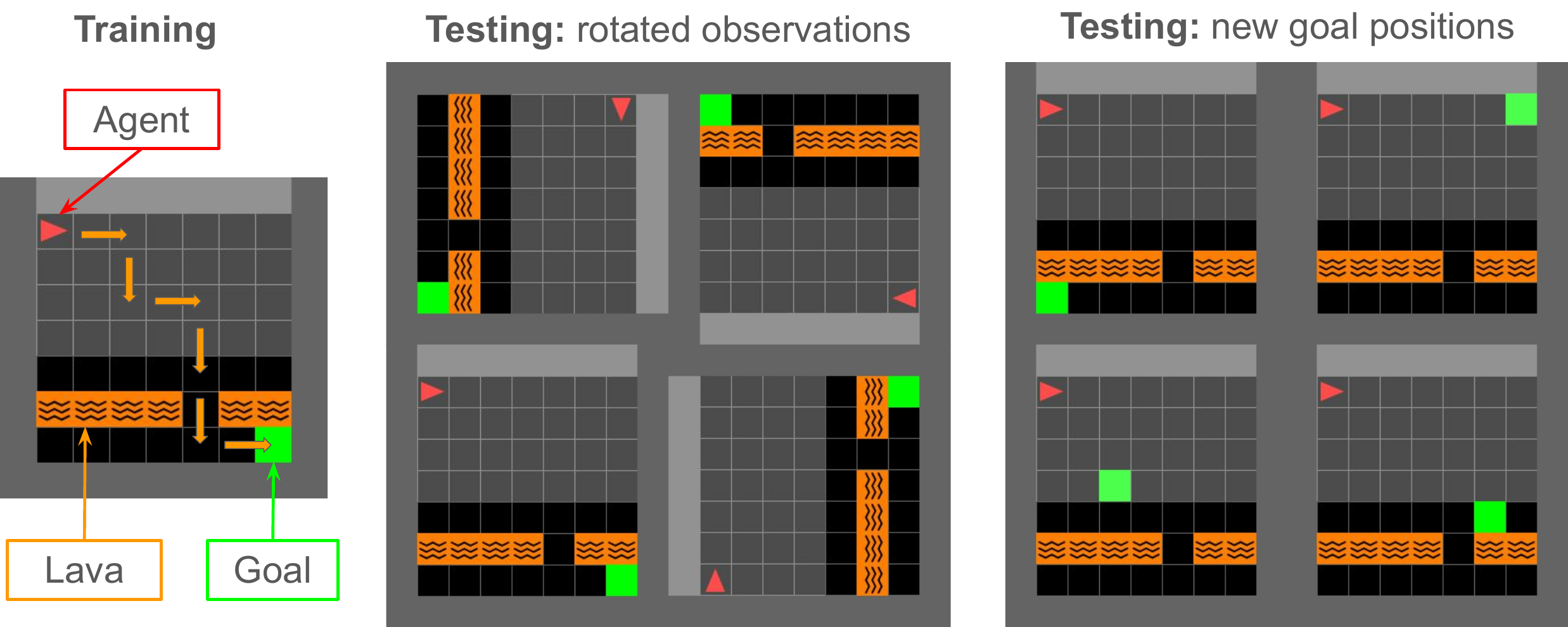}
    \vspace{-0.2 cm}
     \caption{Train vs. test observations of the Mini-grid Lavacrossing (easy-N1) environment. We test generalisation of agents to varying goal and starting positions. 
     }
    \vspace{-0.3cm}
    \label{fig:lavacrossing}
\end{figure}

\textbf{Evaluation \& Results}.The experiments with deep Q-learning~(IMPALA~\citep{espeholt2018impala}) here focuses on GSA with graph matrices ${\color{brown} G_{k,v,q}}$. The number of lava rivers generated in the environment is proportional to the difficulty level. We evaluate the out-of-distribution generalization by training the agent on difficulty level 1 and 2 and testing it on levels 1 to 3 , and varying goals unseen during training (Figure~\ref{fig:lavacrossing}). As shown in Figure \ref{fig:mainres}, SiTs generalise better than the CNNs even on tasks which do not include rotation symmetry. For an ablation see the appendix \ref{app:abl}.



\vspace{-0.15cm}
\subsection{Scaling SiTs: Procgen, Atari 100k \& DM-control}
\label{sec:exp_proc}
\vspace{-0.1cm}

We demonstrate scalability of SiTs in the widely-studied Procgen benchmark~\citep{cobbe20a},  Atari 100k \cite{bellemare13arcade,kaiser2020} and DM-control \cite{DMcontrol}. For details of the latter see the appendix \ref{app:atari} and \ref{app:dmc}, respectively. The Procgen benchmark corresponds to a distribution of partially observable MDPs (POMDPs) $q(m)$, and each level of a game corresponds to a POMDP sampled from that game’s distribution $m \sim q$. The POMDP $m$ is determined by the random seed used to generate the corresponding level. Following the setup from \cite{cobbe20a}, agents are trained on a fixed set of n = 200 levels (generated using seeds from 1 to 200) and tested on the full distribution of levels (generated by sampling seeds uniformly at random from all computer integers). We evaluate test performance on 20 different levels.

\begin{table*}[t]
\centering
\resizebox{2\columnwidth}{!}{
\begin{tabular}{ l |  c | c  c || c c c | c c  || c  }
\toprule
\textbf{Procgen Task} & \textbf{CNN} & \textbf{ \small E2CNN} & \textbf{ \small E2CNN$^{'}$} &\textbf{ViT} & \textbf{SiT} & \textbf{SiT$^*$} & \textbf{SeT} & \textbf{SieT} & {\color{gray}CNN-UCB$^\ast$}~($9\times$) \\
\midrule
CaveFlyer & 4.0\% &  13.4\% &  17.7\% & -1.8\% & \textbf{59.7}\% & \textbf{55.5}\% & 4.6\% & \textbf{34.5}\% & {\color{gray}18.0\%} \\
StarPilot & 36.3\% &  28.1\%   &  29.4\%   & 6.7\% & 31.3\% & 31.0\% & 38.4\% & \textbf{42.2}\% & {\color{gray} 44.6\%} \\
Fruitbot & 70.8\%  &   64.0\%\  & 66.1\%\  & 9.7\% & 69.8\% & 70.5\% & 68.9\% & \textbf{76.0\%} & {\color{gray} 85.8\%} \\
Chaser & 10.6\%  &  13.6\% &  15.6\% & 9.0\% & 35.6\% & 45.6\% & 50.1\% & \textbf{54.0}\% & {\color{gray} 44.6\%} \\
\midrule
Average & 30.4\%  &  29.8 \%  & 32.2 \%  & 5.9  $\%$  & 49.1 $\%$  & \textbf{ 50.6 $\%$}   &  40.5 $\%$   &\textbf{ 51.7 $\%$} & {{\color{gray}48.2$\%$}   }   \\
\bottomrule
\end{tabular}
}
\caption{\textbf{CNN/ViT vs. SiTs on Procgen environments: Caveflyer, Starpilot, Chaser, Fruitbot}. 
We train with PPO (DrAC) + Crop augmentation for SiT (SiT$^\ast$, SeT, SieT) and compare to the CNN,  and E2CNN with Dihedral symmetry group \citep{wang2022mathrmsoequivariant}  (ResNet with model size $79.4k$   comparable to the SiT - $ 65.7k$;  ViT with 4-layers - $ 216k$,
 E2CNN with 4-layers  $ 70.7k$, and  E2CNN${'}$ with 4-layers  $ 139.2k$ (increased features) ). We do not alter the ResNet architecture of \cite{raileanu2020automatic} but chose the same hidden-size of 64 as for the SiT as well reduce the number of channels to [4,8,16]. We train over 25M steps. Following \citet{agarwal2021deep}, we report the min-max normalized score that shows how far we are from maximum achievable performance on each environment. All scores are computed by averaging over both the 4 seeds and over the  23M-25M test-steps. The UCB-DrAC results with Impala-CNN($\times4$) ResNet with 620k parameters are taken from \cite{raileanu2020automatic}.  
}
    \label{tab:main_table1}
    \vspace{-0.4cm}
\end{table*}

 Table~\ref{tab:main_table1}  shows the results for SiT, the equivariant  SeT as well as a both  invariant  $\&$   equivariant SieT trained with PPO (DrAC~\citep{raileanu2020automatic}) with crop-data augmentation.  Sit$^\ast$ uses two  consecutive sums over triangles  of the attention score matrix. 
Invariant SiTs perform well in environments  not reflecting the symmetries of the model, e.g. Starpilot, Fruitbot, Chaser are not rotation invariant. However, the combination of invariance and equivariance of SieTs is superior. 

{\bf Results:} SiT almost doubles the performance on the rotation invariant Caveflyer environment w.r.t. to the ResNet (620k weights) of UCB-DrAC~\citep{raileanu2020automatic}. As UCB-Drac  uses the rotational data-augmentation for training, we can conclude that SiTs outperform rotational data-augmentation. Overall, our tested SiTs, SeT and SieT models ($\approx$ 70k weights ) substantially outperform the CNN and E2CNN (4-layers) \cite{weiler2021general} baselines with similar number of weights  while perform 
comparably to the UCB-ResNet with $9\times$ parameters (620k weights). Notably, all the SiT variants obtain $7-9\times$ improvements in performance compared to ViTs.


{\bf Proof-of-concept Atari 100k and DM-control:} We evaluate our SieT model on 5 common Atari games, SpaceInvaders, Pong, Breakout, KungFuMaster , MsPacman, and find comparable sample-efficiency  to the baseline CNN \cite{hessel2017rainbow, iris2023}, see  Figures~\ref{fig:IRIS2} and \ref{fig:IRIS3} and Table~\ref{tab:atari}. We also test  1D shift-symmetric GSA in the transformer world-model \cite{iris2023}, see Figure~\ref{fig:IRIS}.    DM-control: We employ SieT on top of SAC \cite{SAC} on the Walker-walk task. Without any hyper-parameter and backbone changes compared to our Procgen setup, SieT has comparable performance to the ViT baseline with $>$ 1M weights~\citep{hansen2021softda}. 

{\bf Hyperparameter Sensitivity }. The same limited h-parameter search was performed for SiTs and ViTs. Compared to the ResNet baseline~\citep{raileanu2020automatic}, we employ larger batch-size 96 (instead 8) and PPO-epoch of 2 (instead 3). SiTs don't require  tuning except for the batch-size, e.g., a PPO-epoch of 3 works well too. No tuning at all on DM-control and Atari 100k.
ViTs generally exhibit suboptimal performance in RL, with the notable exception of \cite{hansen2021stabilizing, tang2021the}. We attribute this  that ViTs are less sample-efficient and require different hyperparameter  compared to CNNs. As Vits are  compute extensive hyperparameter search is practically infeasible. {\bf  SiTs are  applicable to RL tasks as they alleviate both of these caveats.}

\begin{figure*}[t!]
 \includegraphics[width=1\linewidth]{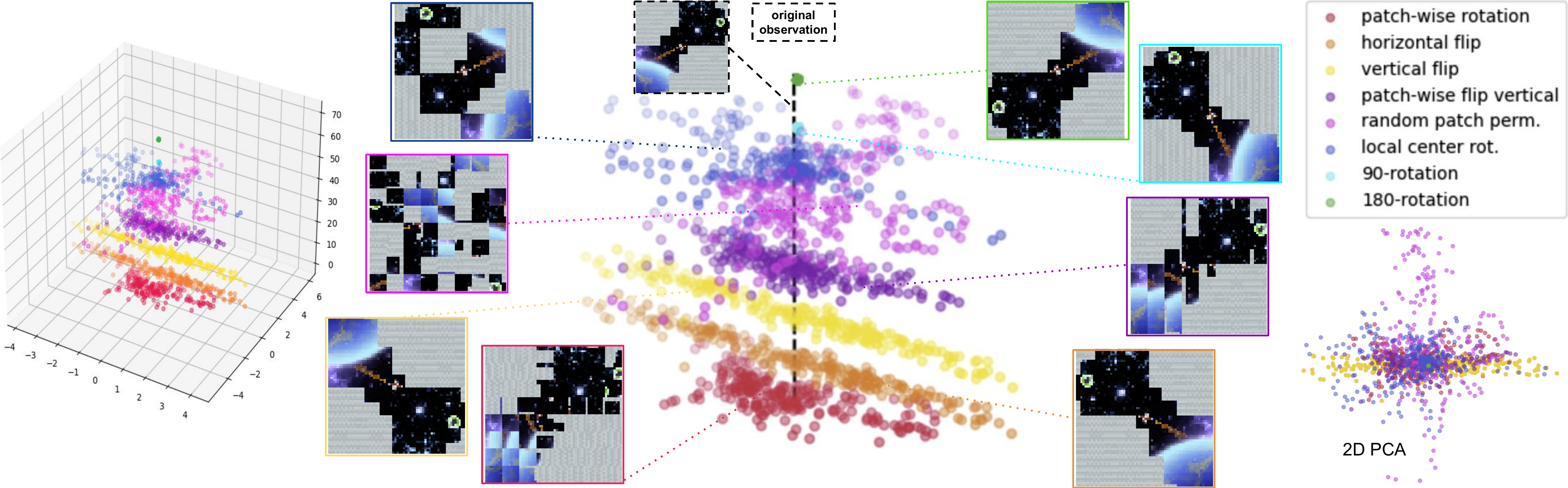}
    \centering
      \caption{PCA of latent space representation of the trained SiT model after 25M steps on Procgen CaveFlyer. We display the  PCA of the difference of the latent representation of augmented observations and original ones. 3D-view  (left and center) - vertical separation for  illustration purposes - 2D-view i.e. from above (bottom-right). }
    \label{fig:PCA}
      \vspace{-0.05cm}
\end{figure*}

{\bf Latent representation analysis:} In Figure~\ref{fig:PCA}, we present a principal component analysis (PCA) of the latent representation of the policy SiT model.  While local symmetries and global flips have been {\bf dynamically broken} during training exact symmetries of {\bf global rotation is preserved}, i.e. all the data-points collapse into (nearly) identical points for the latter. As by design of SiTs, the local attention patch symmetry - {\it local center rot.} in Figure~\ref{fig:PCA} - is broken "softly" as it is relatively close to the original latent representation. Note that at weight initialisation of SiT, all  of the symmetries except permutation invariance are almost exactly preserved.




\vspace{-0.1cm}
\subsection{SiTs beyond RL: Vision Task Ablation}
\vspace{-0.05cm}

We perform an ablation study of SiTs, SieTs compared to  ViTs on a supervised vision task  on the CIFAR-10 dataset \citep{krizhevsky2009learning} see Figure~\ref{fig:gsatt}b; firstly we compare ${\color{mygray}G}$ in SiT to a conventional position embedding in ViT; secondly we use our SieT model with $ {\color{brown} G_{k,q}}, \, {\color{brown} G_{v}}, {\color{mygray}G}=1 $  to show improved sample efficiency \& performance compared to ViTs \citep{vit}.  SiT and SieT   use the  horizontal flip symmetry preserving  graph matrices.

\begin{figure*}[t]
    \centering
\begin{subfigure}{.54\linewidth}
\centering
\includegraphics[width=1\linewidth]{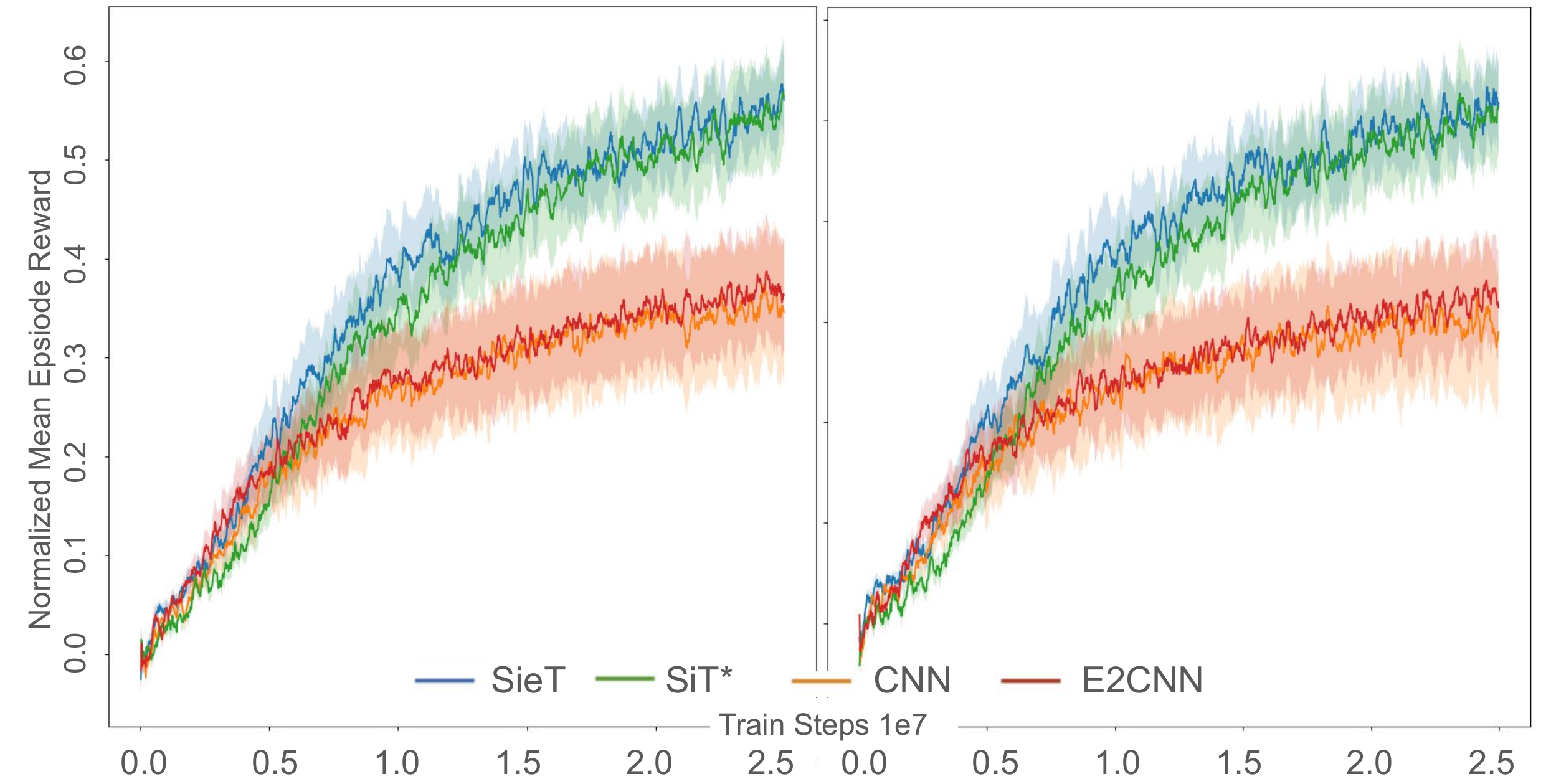}
     \caption{ Train (left) and test curves (right), normalized reward averaged over  the StarPilot, CaveFlyer, Chaser, Fruitbot  environments. }
    \label{fig:imagenet}
\end{subfigure}\hfill
    \begin{subfigure}{.44\linewidth}
\centering
\includegraphics[width=0.8\linewidth]{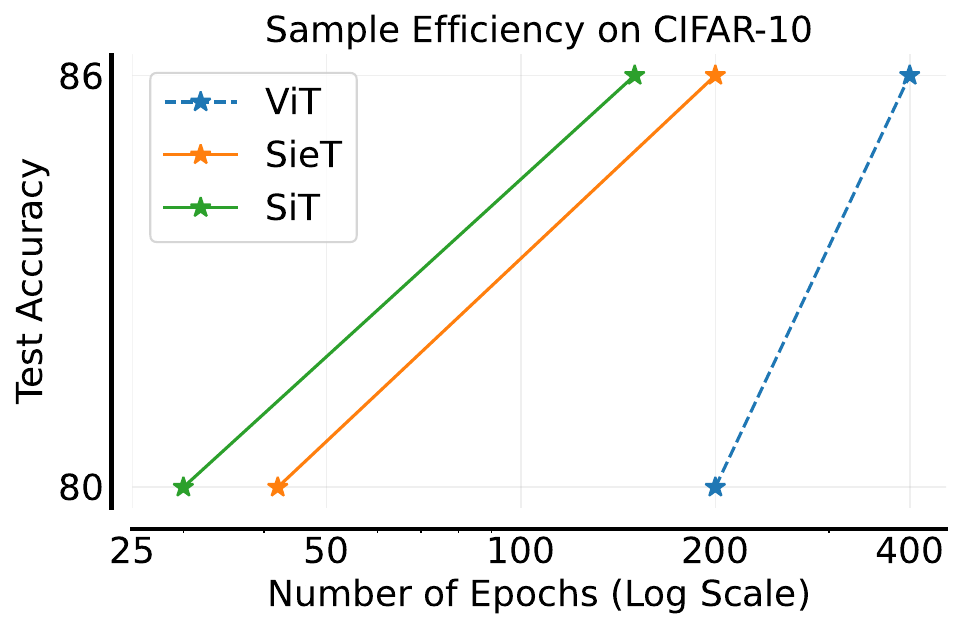}
    \caption{SiTs and SieTs are much more sample efficient than ViTs~\citep{vit}. The SieT model has 4 local GSA layers and 8 global ones.}
    \label{fig:se_cifar}
    \end{subfigure}
    \caption{   SiTs are comparable to CNNs in terms of sample efficiency  on Procgen (a) and outperform conventional ViTs (b).}
    \vspace{-0.15cm}
\end{figure*}

\textbf{Results}. The permutation invariance in  ViTs  is broken by the use of a positional embedding; in SiTs  by the graph matrix ${\color{mygray}G}$ in~Equation~(\ref{eq:defatt3}). While removing positional embedding to obtain local and global permutation invariance substantially degrades performance ~(56\% test accuracy), using our graph symmetric attention (80\% test accuracy) is superior to using positional embedding globally (76\% test accuracy). Furthermore, Figure~\ref{fig:se_cifar} shows that SiTs and SieTs reach same performance as ViTs but using $2-5\times$ less training epochs. 
We perform extensive ablations studies on the impact of patch-size changes of SiTs on CIFAR 10 in appendix \ref{app:ps-sit}.

\subsection{Entire Procgen Suite SieT}
\label{sec:tinyProc}
\begin{figure*}[t]
    \centering
\begin{subfigure}{.54\linewidth}
\centering
\includegraphics[width=0.8\linewidth]{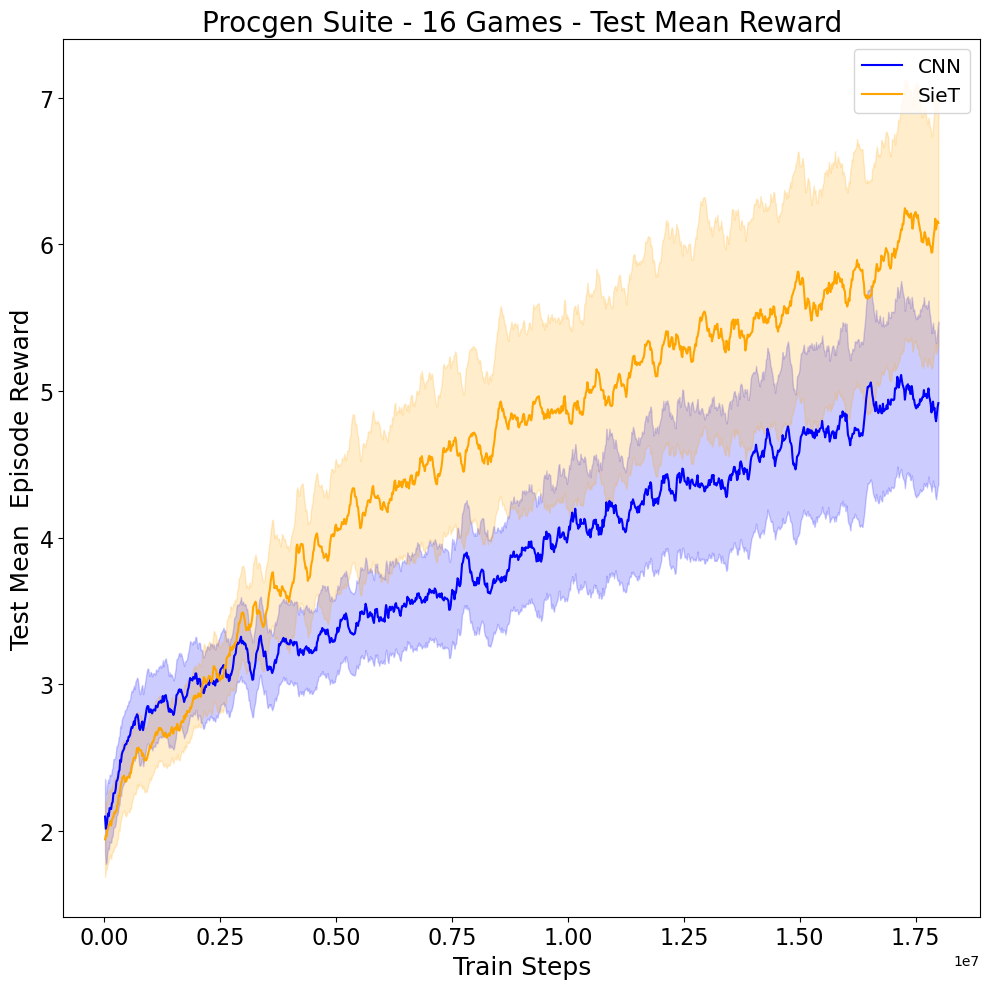}
     \caption{Mean episode  test  reward averaged over 16 games, and 4 seeds, respectively.  }
    \label{fig:imagenet}
\end{subfigure}\hfill
    \begin{subfigure}{.44\linewidth}
\centering
\includegraphics[width=1.0\linewidth]{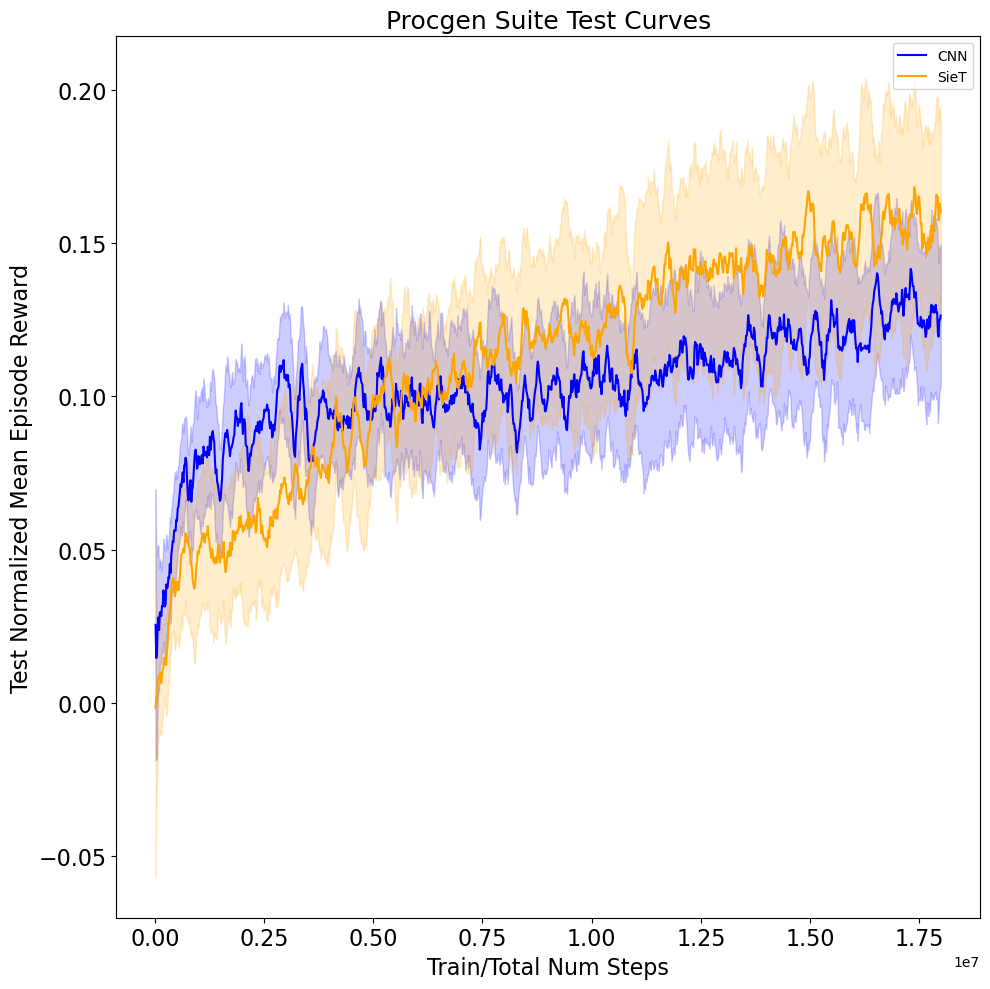}
    \caption{Normalized mean episode  test  reward averaged over 16 games, and 4 seeds, respectively. }

    \end{subfigure}
    \caption{ Empirical evaluation on entire Procgen suite i.e the reward is averaged over 16 games and standard error of  SieT model  (~41k weights) and a CNN (~40k weights) on top of DrAC with crop augmentation over 18M training steps. The CNN consist of 4 convolutions layers with channel dimensions $[4,8,16,32]$, respectively. The SieT model contains two local GSA layers with features $[16,32]$, and to global ones with 64 features, respectively. both the CNN and SieT policy have  a hidden-size of 64.}
    \label{fig:P16}
\end{figure*}

In this section we evaluate a smaller SieT variant (~41k weights) on the entirety of the Procgen suite of 16 games, see Figure~(\ref{fig:P16}). We conclude that SieT outperforms a comparable CNN on average over the 16 games. The lower total score compared to Table~(\ref{tab:main_table1}) is due to two main factors. First, both the tiny SieT and CNN variant do not perform on several environments in combination with DrAC. Second, the game selection in Table~(\ref{tab:main_table1}) reflects a collection of very successful environments of DrAC.

\subsection{Limitations of SiTs}

Equivariant CNN based  networks \citep{wang2022mathrmsoequivariant}  have shown promise in particular in environments which admit symmetry but also where latent dynamics admit symmetries \cite{wang2023the}. At the core we provide a Transformer based alternative which outperforms E2CNN for comparable model sizes see Table~(\ref{tab:main_table1}). Thus in general Sit will offer benefits over CNNs in environments where at least some remnant symmetries other than shift symmetries are present - this is in accordance with our findings. 
Additionally, we observed benefits of SiTs in terms of their generalization performance due to unseen tasks, while  performance in single-task RL on the same training and test task settings seem to be comparable to CNNs. 
It is worth noting that except for the Caveflyer environment, other environments in ProcGen do not admit strong symmetry properties, which would favor SiTs over CNNs off the shelf.

From our ablation studies on Minigrid see appendix Table~(\ref{tab:table_ablation}) we conclude that local GSA is of significant importance to get the performance benefits. However, as we discuss in section~\ref{sec:scaleSit4} “Scalability of Sit” the local GSA layer poses compute and memory challenges for application to larger model sizes  and complex tasks. 

Let us stress  that we have evaluated  SiTs (and SieTs) on a large variety of tasks, namely Minigrid, Procgen, Atari, DM-control, and CIFAR10. The major limitaion to Sits currently are the compute and memory expenses that arise in the SiT architecture  from  the local GSA. With increased  computational cost for each {\bf local GSA} layer added. The latter,  trade-offs for improved sample-efficiency see Figure~(\ref{fig:se_cifar}).  Our work serves as a proof-of-concept for the approach and work while more efficient future custom implementation of SiTts may alleviate some of the computational cost.

One specific  example of failure is Procgen  in the new rebuttal study see section \ref{sec:tinyProc} is that both the  SieT and CNN variant do not learn the BossFight task very well. However, on the CNN side when adding more convolutional layers adequately the model starts to learn well.

\vspace{-0.1cm}
\section{Related Work}
\vspace{-0.1cm}
Symmetry is a prevalent implicit approach in deep learning for designing neural networks with established equivariances and invariances. The literature on symmetries in Vision Transformers~(ViTs) \cite{NEURIPS2020_15231a7c,romero2021group} is relatively limited compared to CNNs~\citep{Zhang1988, LeCun1989, Zhang1990}, recurrent neural networks \citep{Rumelhart1986, Hochreiter1997}, graph neural networks \citep{Maron2019, Satorras2021}, and capsule networks \citep{Sabour2017}. PI in  attention mechanisms and  ViTs has been examined in \citep{pmlr-v97-lee19d} and \citep{tang2021the}. In contrast, the SiT variants  admit different adaptive symmetries other than PI.  

The Region ViT method \citep{chen2022regionvit} divides the feature map into local areas, where each region has tokens that attend to their local counterparts. 
We use global tokens and use local attention in a neighbouring subset. The method in \citep{wang2021crossformer} combines local and global attention to reduce complexity, focusing globally on specific windows. For us "global" means standard attention, while "local" pertains to attention within a window. 


 Conventionally  sample efficiency is enhanced by data augmentation \citep{Krizhevsky2012}. Simple image augmentations, such as random crop \citep{NEURIPS2020_e615c82a} or shift \citep{Yarats2021}, can improve RL generalisation performance;  in particular when combined with  contrastive learning~\citep{agarwal2021contrastive}. 
SiTs are complementary to data-augmentation.

Algebraic symmetries in Markov Decision Processes (MDP) were initially discussed in \citep{homo} and recently contextualized within RL in \citep{vanderPol}.
Symmetry-based representation learning \citep{higgins}  refers to the study of symmetries of the environment manifested in the latent representation and was extended to  environmental interactions in \cite{NEURIPS2019_36e729ec}. These concepts were recently extended in \cite{rezaeishoshtari2022continuous,pmlr-v162-mondal22a}. In \cite{weissenbacher2022koopman}, symmetries of the dynamics are inferred in a self-supervised manner; \cite{cheng2023look} discusses time-reversal symmetry. These approaches are mostly complimentary to employing  SiTs with  equivariance/invariance  which may aid the former.

Numerous prior works have demonstrated the exceptional sample efficiency of RL achieved through equivariant methods with CNNs \citep{vanderPol2019, wang2022so}. 
 Steerable Equivariant CNNs named E2CNNs \cite{pmlr-v48-cohenc16,weiler2021general} have been widely applied to RL \cite{,mondal2020group,wang2022mathrmsoequivariant}. In contrast, SiTs belong to ViT paradigm, i.e. a distinct new approach to achieve invariance as well as equivariance both locally and globally.  
Approximately equivariant networks \citep{pmlr-v162-wang22aa}  offer a flexible and adaptive approach by imposing constraints on the weights via a regularizer.   Rotation  invariance steerable convolution in toy-examples is discussed in  \cite{zhao2023integrating}.  SiTs start from a manifest PI  and describe an implicit adaptive mechanism of breaking it and scale to relevant tasks in RL. 

\vspace{-0.1 cm}
\section{Conclusions}
\vspace{-0.1cm}
In this work, we introduced the Graph Symmetric Attention (GSA) mechanism, a symmetry-preserving attention layer that adapts the self-attention mechanism to maintain graph symmetries. We combine GSA with ViTs to propose the novel SiT architecture. By leveraging the interplay of local and global information, SiT achieves inherent out-of-distribution generalization on RL environments.

Transformers have significantly advanced natural-language-processing and vision tasks, particularly in scalability. Our work may pave the way for applying these benefits to image-based RL. Additionally, transformers facilitate integration with Large Language Models (LLMs) for multimodal architectures, highlighting their potential in future vision and language-based RL research. 

\section*{Impact Statement}
This paper presents work whose goal is to advance the field of Machine Learning. There are many potential societal consequences of our work, none which we feel must be specifically highlighted here.

\section*{Acknowledgments}
We would like express our gratitude to Yujin Tang for helpful comments on the draft. 
And, we would like to thank Y.\, Nishimura for technical support. This work was supported by JSPS KAKENHI (Grant Number JP22H00516), and JST CREST (Grant Number JPMJCR1913).




\bibliography{iclr2024_conference}
\bibliographystyle{icml2024}


\newpage
\appendix

\section{Overview of Definitions:  Graph Symmetric Attention Mechanism}
\label{app:details}

 \begin{figure*}[t!]
  \vspace{-0.45cm}
 \includegraphics[width=\linewidth]{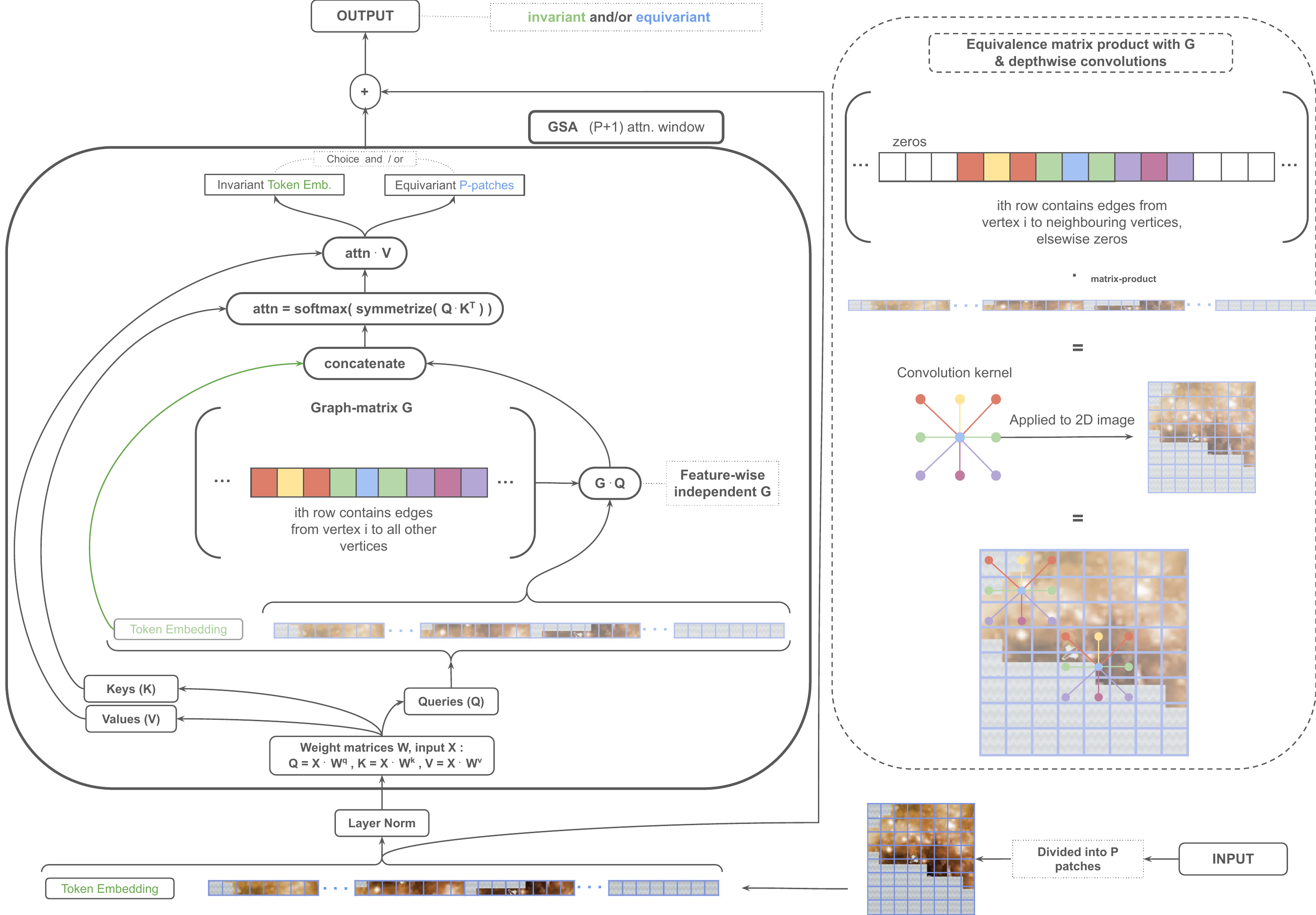}
    \centering
     \vspace{-0.45cm}
      \caption{GSA module architecture (left), and equivalence of graph matrix  and depth-wise convolutions (right). The GSA layer includes Layer-Norm and additive skip connection in addition to the GSA module. We display the GSA variant ${\color{brown} G_{k,v,q}}$ with ${\color{brown} G_{k,v}} =1 $. }
    \label{fig:SitArchitecture}
\end{figure*}

We propose the following  {\bf G}raph {\bf S}ymmetric {\bf A}ttention (GSA) mechanism in \eqref{eq:defatt3app}.
 \begin{figure*}
\bea\label{eq:defatt3app}
\text{GSA}(K,V,Q) &=&  \text{softmax}  \Big(\, \tfrac{1}{\sqrt{d_q}}\,  \Gamma(Q,K) \,  \big)\, \, {\color{brown} G_v} \, \,\, \, V \;\;\\[0.2cm] \nonumber \text{with} \;\; \Gamma(Q,K) &=& \text{symmetric} \Big( \,{\color{blue} \sigma\, \Big( }  {\color{blue} G_{qk}}  \, \,\, \,\big( \; {\color{brown}  G_q}  \,\,  Q  \,\, \,\, \,\, [ \; {\color{brown}  G_k}   \,\, K \;]^T\, \big) \,\,  + \, {\color{blue}G_b}   {\color{blue}\Big)} \, \odot \, {\color{mygray}G} \Big)\;\;.
\eea
\end{figure*}
Where  $ {\color{mygray}G}, \, {\color{brown} G_{k,v,q}} , \,{\color{blue} G_{kq , b}} \,\in \, \mathbb{R}^{P\times P}$ being the graph matrices described in figure \ref{fig:gsatt} and $\sigma$ is an activation function . The color coding refers to different conceptual implementations, which may be used in combination. When using token embeddings $K,\,
 V,\, Q,\, \in \, \mathbb{R}^{P+1 \times d_{f}}$ thus $ {\color{brown} G_{k,v,q}} \,\in \, \mathbb{R}^{P+1 \times P+1 \times d_f}$ and $ {\color{gray}G},\,{\color{blue} G_{kq , b}} \,\, \in \, \mathbb{R}^{P+1 \times P+1 \times \#heads}$. In particular, that implies that we apply a different set of graph weights to the  feature and head dimensions. See Figure~(\ref{fig:SitArchitecture}) for a visualisation.
 

Moreover, we propose the a second variant of  the GSA$^{a-sym}$ of our attention mechanism which replaces the attention matrix as  
\bea\label{eq:defatt4}
 \text{softmax}  \Big(\, \tfrac{1}{\sqrt{d_q}}\, \text{sym} \big(\,  \Gamma(Q,K)\, \big) \Big) \,\\[0.2cm] \nonumber 
 + \,\, \text{softmax}  \Big(\,\tfrac{1}{\sqrt{d_q}}\, \text{a-sym} \big( \,\Gamma(Q,K) \,\big)  \,  \Big)
\eea
where we symmetrise and anti-symmetrises over the patch indices of $\Gamma(Q,K)$ respectively, the latter is as in \eqref{eq:defatt3}; and {\it a-sym} refers to replacing the symmetrisation in \eqref{eq:defatt3app} by anti-symmetrisation. Anti-symmetrisation of a matrix M refers to $M_{ij} \to M_{ij}-M_{ji}$.

{\bf Empirical Evaluation Omission Overview:} 
\begin{itemize}
    \item On our grid-world environment experiments we found qualitatively that the variants $ {\color{blue} G_{kq , b}}, \, {\color{mygray}G} $   required more h-parameter tuning to show comparable performance to the CNN baseline. We thus removed a  quantitative analysis from the paper.\footnote{  Let us stress that both \( \color{mygray} G \) and \(\color{mybrown} G_{v,k,q} \) are indeed sufficient to break symmetries, respectively, i.e. to achieve the desired equivariance.}
    \item Adding the anti-symmetrisation \eqref{eq:defatt4}   to the architecture increased generalisation performance on the grid-world environment. However, it requires two softmax operation, which renders it hard to scale; we thus removed a  quantitative analysis from the paper.
\end{itemize}

\textbf{More formally of can define e.g. G in Figure \ref{fig:gsatt} (c).} Pick  $G  \in \mathbb{R}^{P \times P}$ such that it admits a shared weight if the distance between vertices is the same. For more technical details see the appendix \ref{app:details}
More formally, $G_{ij} = \theta^{(\kappa)} \, , \,{\tiny i,j\, =\, 1,..,P}$ with weights $\theta$ with labels $\kappa =1,..,{\small \#(\text{unique edge lengths of 2D grid graph})}$. The element $G_{ij}$ corresponds to the edge between the $i^{th}$ and $j^{th}$ vertex, i.e.\,the assigned weight index $\kappa$ is identical if and only if the distance between the $i^{th}$ and $j^{th}$ vertex is the same.

\paragraph{Rotational Symmetry.} To ensure that the layer solely possesses rotational symmetry, it is essential to disrupt the flip symmetry. This can be accomplished by selecting flip and rotational graph matrices, as depicted in fig. \ref{fig:gsatt} (c), and summing over distinct directed subgraphs with three vertices, i.e., triangles, while assigning weights to each contribution
see \eqref{eq:rot_app}.
\begin{figure*}[t]
 \bea
 \label{eq:rot_app}
 \text{GSA}_{\text{rot}}(K,V,Q) & = & \text{softmax}  \Big(\, \tfrac{1}{\sqrt{d_q}}\,  \sum_{\text{ triangle edge = 1}}^{3 \, P^2} \, \Theta_{\text{tri.\,edge}} \, \Gamma(Q,K)_{\text{tri.\,edge}} \,  \Big)\, \,V \  \\[0.2cm] \nonumber
 &=& \, \text{softmax}  \Big(\, \tfrac{1}{\sqrt{d_q}}\,  \Gamma^{\text{rot}}(Q,K) \Big)\, V  \;\;, \,\, \text{ where}
 \eea
 \end{figure*}
 \begin{figure*}[t]
  \beq
  \Gamma^{\text{rot}}(Q,K)_{ij} \;\; =\;\;  \Theta^{(i\to j \to k)}\, \Gamma(Q,K)_{ij}
  \, + \,\Theta^{(j\to k \to i)} \,\Gamma(Q,K)_{jk} \, + \, \Theta^{(k\to i \to j)}\, \Gamma(Q,K)_{ki} \;\; ,
 \eeq
  \end{figure*}
 
In essence, this implies that for any component of $\Gamma(Q,K)_{\text{edge}}$, two additional entries are added, all weighted with $\Theta's$.
The $\Theta's$ are trainable parameters, which will be shared if the angle between two edges of the triangle is identical.

 In \eqref{eq:rot}, the third vertex of the triangle is chosen as a function of $i,j \mapsto k = T(i,j)$ for a unique map  $T$; the  weights $\tiny \Theta's$ are shared if the angle between edges of the triangle in the square grid is identical.  The label $\tiny {(i\to j \to k)}$ denotes the angle at the $\tiny j^{th}$  vertex i.e. between the $(ij)$ and $(jk)$ edge. 
In essence, this implies that for any of the $P^2$-components of $\Gamma(Q,K)$, two additional entries are added, all weighted with $\Theta's$.

A unique triangulation of the square grid is be chosen as follows.
Disregarding the heads-dimension for the time being, any entry of the matrix $\Gamma(Q,K)$ can be construed as a connection between a specific patch and another, thereby enabling the drawing of an edge from the former to the latter. This forms the initial directed edge of the triangle, linking two vertices. From the end of the latter, we opt to turn right and proceed to the nearest vertex in the grid, constituting the third vertex of the triangle. The direction of the edges is determined by traversing the triangle. Notably, as flips modify left/right and/or up/down, the aforementioned sum is not invariant under flips; however, it preserves left/right rotations, which transform the grid into itself.


{\bf Duality to convolutional kernels.} See Figure~(\ref{fig:SitArchitecture}) and (\ref{fig:gsatt}) for a visualisation. The shared nature of the graph-matrix elements make them identical to a 2D-convolutional layer if the $kernel \,size \,\, = \,\, 2 \,*\, image \, size +1$, for a square image. If a  specific kernel size is chosen it corresponds to the graph matrix with zero entries for vertices of larger distance than the kernel size, see Figure~(\ref{fig:SitArchitecture}).

\subsection{One-dimensional Data / Sequential Data}
\label{secapp:1D}

 While our discussion  centers on the 2D grid, GSA may be adapted to 1D data where it ensures shift-symmetry and an optional flip-symmetry.

One may define G formally for the 1D case as.
The data points in the 1D data are consider vertices, then assume that the distances are measured w.r.t. a specific vertex, and edges can be viewed as vectors. Then, pick $G \in \mathbb{R}^{P \times P}$ such that a shared weight is present in $G$:
\begin{enumerate}
\item  When horizontal component of edges have the same magnitude ( flip-preserving)
\item  When the magnitude of the edges is same, but direction left/right i.e. the sign is accounted for which results in shift-symmetry.
\end{enumerate}
Points  (1) and (2) above are completely analog to the more complicated 2D case which is proofed in section~\ref{app:proof}. Flip-preserving  1D GSA (1) for sequential data implements  time-reversal symmetry, recently  discussed in the context of RL \cite{cheng2023look}. 


{\bf Application of Transformers in RL on Sequence data:} The rise of Transformers in model-based RL \cite{chen2021decision,iris2023} opens up another direction which may adapt our approach.

\section{Overview: Symmetry-invariant Transformer}
 \begin{figure*}[t!]
 \includegraphics[width=0.80\linewidth]{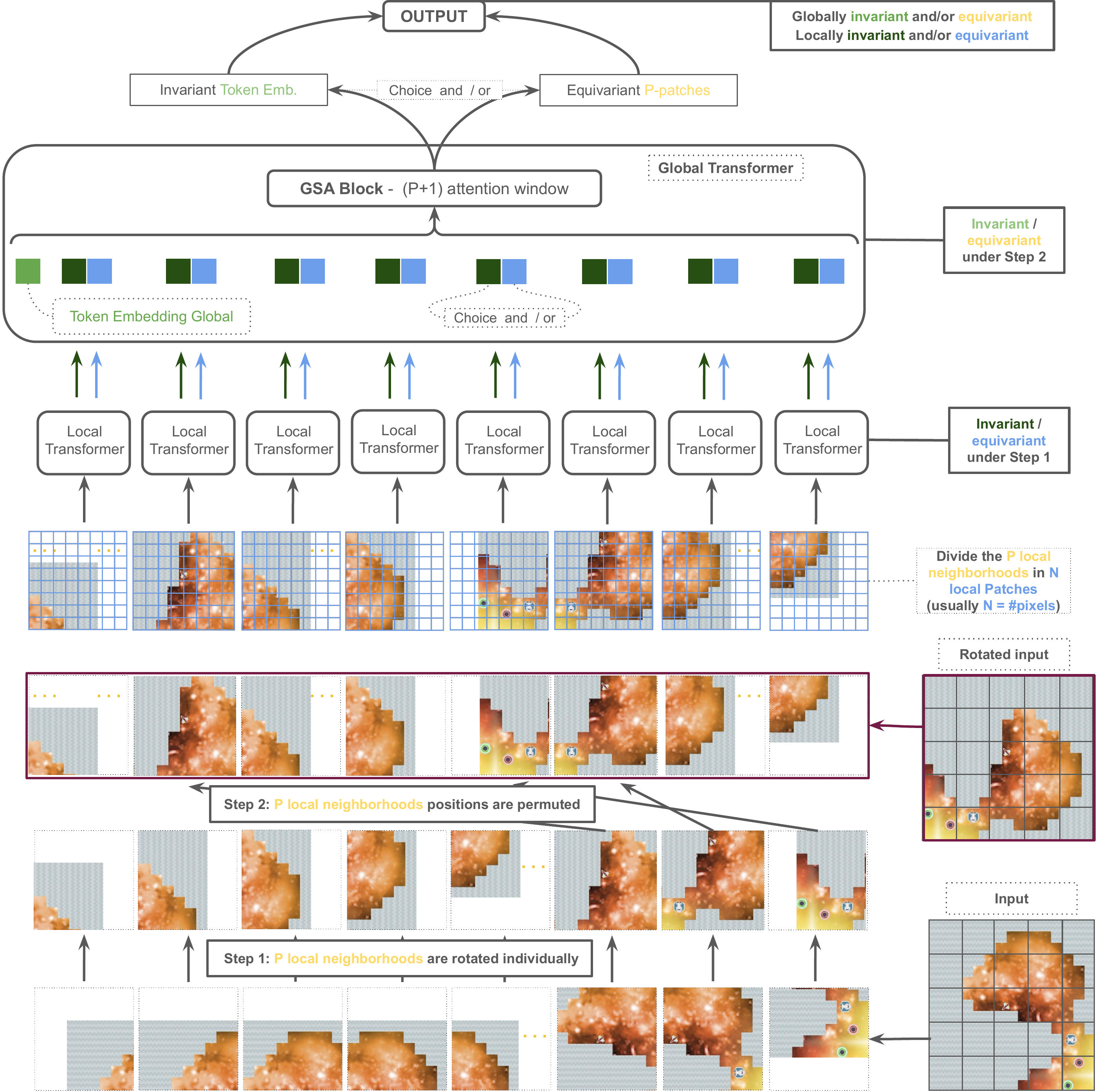}
    \centering
      \caption{ Illustration on exact global symmetries of SiTs;  obtained as a result of the combined effect of local and global GSA modules. }
    \label{fig:SitSym}
\end{figure*}

See Figure~(\ref{fig:SitArchitecture_main}) for a schematic overview of the SiT architecture and for details on the symmetry of SiTs Figure~\ref{fig:SitSym}. The local attention window's are processed by a GSA, which then passes only the token embedding dimension onto the global GSA layers. Since the token embedding dimension is invariant under the respective symmetries the global GSA receives a symmetry-invariant input of the local patches.

\lstdefinestyle{mystyle}{
    basicstyle=\small\ttfamily,
    breakatwhitespace=false,
    breaklines=true,
    captionpos=b,
    keepspaces=true,
    showspaces=false,
    showstringspaces=false,
    showtabs=false,
    tabsize=2,
    language=Python,
    morekeywords={qkv_SymBreak, self.G_weights,  self.G_idxs, graph_function},
    keywordstyle=\color{blue},
    escapeinside={(*@}{@*)}
}

\section{Algorithm details }
\begin{figure*}[t!]
\begin{lstlisting}[style = mystyle,language=Python, caption= Pseudocode for GSA (PyTorch-like). Changes relative to self-attention in \color{brown}{brown}., breaklines=true  morekeywords={qkv_SymBreak, self.G_weights,  self.G_idxs, graph_function}, keywordstyle=\color{blue}] 
class GraphSymmetricAttention(Module):
    def __init__(self, dim, num_patches, num_heads):
        self.num_heads = num_heads
        self.qkv = Linear(dim, dim * 3) # Fully-connected layer
        # Generate shareable graph weights 
        (*@\color{brown}self.G\_weights, self.G\_idxs  = graph\_function(num\_patches, dim * 3)@*)

    def  (*@\color{brown}qkv\_SymBreak(self,x):@*) 
        # Select indices of weights along dimension
        (*@\color{brown}G =  index\_select(self.G\_weights , 1, self.G\_idxs)@*)
        # Matrix mulitply with G; omit token embedding dimension
        (*@\color{brown}y = G@x[:,1:,:]@*)
        (*@\color{brown}return cat([x[:,0:1,:], y],dim=1)@*)
    
    def forward(self, x):
        B, N, C = x.shape
        qkv = (*@\color{brown}{self.qkv\_SymBreak}@*)(self.qkv(x))
        qkv = qkv.reshape(B, N, 3, self.num_heads, C // self.num_heads).permute(2, 0, 3, 1, 4)
        q, k, v = qkv[0], qkv[1], qkv[2]
        attn = (q @ k.transpose(-2, -1)) 
        # symmetric attenion (if used)
        (*@\color{brown}attn = attn +  attn.transpose(-2, -1)@*) 
        attn = attn.softmax(dim=-1)
        return (attn @ v ).transpose(1, 2).reshape(B, N, C)
\end{lstlisting}
\end{figure*}

\begin{figure*}[t!]
\begin{lstlisting}[style = mystyle,language=Python, caption= Pseudocode for shared weight graph indices (PyTorch-like)., breaklines=true  morekeywords={qkv_SymBreak, self.G_weights,  self.G_idxs, graph_function}, keywordstyle=\color{blue}]
def (*@\color{brown}graph\_function@*)(num_patches, in_features):
    #compute simple ditance of vertices in square grid
    for k in range(0,num_patches):
        for l in range(0,num_patches):
            for i in range(0,num_patches):
                for j in range(0,num_patches):
                    I = k*num_patches+l
                    J = i*num_patches+j
                    dist[I,J] =  sqrt((k-i)**2 + (l-j)**2)

    #identify equal ditances and associate unique index 
    unique = dist.flatten().unique()
    dim = unique.shape[0]
    for i in range(dim):
        mask = (dist == unique[i])
        idxs[mask] = i
    #initialise independent weights
    weights = Parameter(Tensor(in_features,dim))  
    return weights , idxs
\end{lstlisting}
\end{figure*}

The implementation of the rotation symmetry breaking is a bit lengthy however straightforward. We omit details of how to compute the angles of triangles of vertices on the square grid, and just assert the function "rot\_grid\_to\_idx" here.

\begin{figure*}[t!]
\begin{lstlisting}[style = mystyle,language=Python, caption= Pseudocode for GSA with rotation symmetry(PyTorch-like). Changes relative to self-attention in \color{brown}{brown}., breaklines=true  morekeywords={qkv_SymBreak, self.G_weights,  self.G_idxs, graph_function}, keywordstyle=\color{blue}]
class GraphSymmetricAttention(Module):
    def __init__(self, dim, num_patches, num_heads):
        (*@\color{brown}{self.rot\_SymBreak = Rotation\_Symmetry( num\_patches, num\_heads)}@*)
   
    # ... we only highlight the differnces in the forward pass
    
    def forward(self, x):
        B, N, C = x.shape
        qkv = (*@\color{brown}{self.qkv\_SymBreak}@*)(self.qkv(x))
        qkv = qkv.reshape(B, N, 3, self.num_heads, C // self.num_heads).permute(2, 0, 3, 1, 4)
        q, k, v = qkv[0], qkv[1], qkv[2]
        attn = (q @ k.transpose(-2, -1)) 
        # symmetric attenion (if used)
        (*@\color{brown}attn = attn +  attn.transpose(-2, -1)@*) 
        attn =  (*@\color{brown}{self.rot\_SymBreak}@*)(attn)
        attn = attn.softmax(dim=-1)
        return (attn @ v ).transpose(1, 2).reshape(B, N, C)

class (*@\color{brown}{ Rotation\_Symmetry}@*)(Module):
    def __init__(self, num_patches, num_heads):
        #compute unique angles of triangles in square grid
        idxs, idxs_angles, dim = rot_grid_to_idx(num_patches)
        self.idxs = idxs
        self.idxs_angles = idxs_angles 
        #initialise independent weights
        self.angles =  Parameter(Tensor(num_heads,dim))  
            
    def forward(self, x):
        bs, heads, ps, ps= x.size()
        #index only non-token dimensions
        y = x[:,:,1:,1:].flatten(-2)[:,:,self.idxs]
        #preserve rotation/translation invariances, break flip invariances
        angles = index_select(self.angles,1,self.idxs_angles.flatten())
        #multply vertex with corresponding weight and sum over triangle
        y = (y*angles).reshape((bs, heads, ps-1, ps-1,3)).sum(-1)
        #attach token dimension 
        y = cat([ x[:,:,:,0:1],cat([x[:,:,0:1,1:], y],dim=2)],dim=-1)
        return   y.reshape(x.shape)
        


\end{lstlisting}
\end{figure*}

Lastly, in order to achieve scalability of SiTs we need to establish a connecting between graph matrices and depth-wise convolutions with graph-weights as kernels. The latter implementation is significantly more memory efficient. We note here that the below code  is equivalent to  multiplication of  graph matrix given  in Lisitng 1.

\begin{figure*}[t!]
\begin{lstlisting}[style = mystyle,language=Python, caption= Pseudocode for an efficient GSA (PyTorch-like). Changes relative to Listing (1) and (2) are presented., breaklines=true  morekeywords={qkv_SymBreak, self.G_weights,  self.G_idxs, graph_function}, keywordstyle=\color{blue}]

def (*@\color{brown}graph\_function@*)(kernel_size):
        dist_Mat = torch.zeros((kernel_size,kernel_size))

        #only works correcttly if kernel_size is odd
        i0,j0 = (kernel_size -1) //2, (kernel_size -1) //2
        for i in range(0,kernel_size):
            for j in range(0,kernel_size):
                        distance = math.sqrt((i0-i)**2 + (j0-j)**2)
                        dist_Mat[i,j] =  distance
                        
        
        unique = torch.unique(dist_Mat)
        idxs = torch.zeros(dist.shape)
        dim= unique.shape[0]
        for i in range(dim):
            mask = (dist == unique[i])
            idxs[mask] = i
        
        #initialise independent weights
        weights = Parameter(Tensor(in_features,dim))  
        return weights , idxs 

class GraphSymmetricAttentionEfficient(Module):
    ...
    def  (*@\color{brown}qkv\_SymBreak(self,x):@*) 
        # Select indices of weights 
        #of shape (in\_features,1, kernel\_size, kernel\_size)
        (*@\color{brown}G =  index\_select(self.G\_weights , 1, self.G\_idxs)@*)
        # depthwise convlution with weights G; omit token embedding dimension
        (*@\color{brown}y =  conv2d(x, G, padding= (kernel\_size-1)//2, groups=in\_features)@*)
        (*@\color{brown}return y @*)
   ...
\end{lstlisting}
\end{figure*}

\section{GSA - Graph Symmetric Attention }
In the following  \( K \), \( V \), and \( Q \) denote the keys, values, and queries respectively. They are derived from the input \( X \):
$ K = X W^k, \, V = X W^v, \, Q = X W^q$,
where \( W^q \), \( W^k \), and \( W^v \) are the corresponding weight matrices, i.e. $ V_{ia} = \sum_{=1}^{d_{f}}  X_{i x} W^{v}_{x a}$
The  permutation invariant self-attention layer \citep{pmlr-v97-lee19d}  is given by
\bea\label{eq:defattapp}
Att(K,V,Q) &=& f(Q,K) \, \,\, \, X \, \,\, \, W^v \;\;\\[0.2cm] \nonumber \text{with} \;\; f(Q,K) &=& \tfrac{1}{\sqrt{d_q}} \text{softmax} \left( Q \, \,\, \, [ X \, \,\, \, W^k ]^T  \right)\;\;,
\eea
which can be rewritten using explicit  indices.  One finds
\bea \nonumber
Att(K,V,Q)_{q\,a} &=&  \sum_{i=1}^{P}\,\;\;  \text{softmax} \Big(\,\tfrac{1}{\sqrt{d_q}}\,  f(Q,K)_{q\,i} \Big) \, V_{i\,a}  \;\;\\[0.05cm]  \label{eq:defattapp2a}
\text{with} \;\; f(Q,K)_{q\,i} &=& \sum_{a=1}^{\# heads} \;\; Q_{q \,a} [K]^T_{\, a \, i}  \;\;,
\eea
  For the original self-attention mechanism \cite{vit} one simply needs to use a non-fixed $Q = X\, \,\, \, W^q$  in \eqref{eq:defattapp2a} as 
\bea\nonumber
Att(K,V,Q)_{q\,a} &=&  \sum_{i=1}^{P}\,\; \tfrac{1}{\sqrt{d_q}} \text{softmax} \Big(\; f(Q,K)_{q\,i} \, \Big) \, V_{i\,a}  \;\;\\[0.05cm]  \nonumber
\text{with} \;\; f(Q,K)_{q\,i} &=&  \sum_{a=1}^{\# heads} \;\; Q_{q \,a} \,[\, K \,]^T_{\, a \, i}  \;\;,
\eea
We propose the following three variants of {\bf G}raph {\bf S}ymmetric {\bf A}ttention (GSAention) layer - all of which separately preserves the identical symmetries -
\bea
\text{GSA}(K,V,Q) &=&  \text{softmax}  \Big(\, \tfrac{1}{\sqrt{d_q}}\,\Gamma(Q,K) \,  \Big)\, \, {\color{brown} G_v} \, \,V \, \nonumber \\[0.2cm]  \text{with} \, \Gamma(Q,K) &=&  \,\text{symmetric}  \,\Big({\color{brown}  G_q}  \,\ Q \, \, [  {\color{brown}  G_k}   \, K  \,]^T\, \Big) \,,\,\,\label{eq:defatt3app}
\eea
and
\bea\label{eq:defatt3app2}
\text{GSA}(K,V,Q) &=& \text{softmax}  \Big(\, \tfrac{1}{\sqrt{d_q}}\, \Gamma(Q,K) \,  \Big)\,  \, V \, \\[0.2cm] \nonumber \text{with}  \Gamma(Q,K) &=& \,\text{symmetric}  \, {\color{blue} \sigma\, \Big( }  {\color{blue} G_{qk}} \,\big( Q   \,[  K  ]^T\, \big)   + \, {\color{blue}G_b}   {\color{blue}\Big)} \, ,
\eea
and moreover
\bea\nonumber
\text{GSA}(K,V,Q) &=&  \text{softmax}  \Big(\, \tfrac{1}{\sqrt{d_q}} \, \Gamma(Q,K) \,  \Big)\, \, \,\, \, V \, \,\, \, W^v \;\;\\[0.2cm] \nonumber \text{with} \;\; \Gamma(Q,K) &=& \,\text{symmetric}   \,\Big(  Q \, \, \,\, \, [ \,  K  \, ]^T\, \Big) \, \odot \, {\color{mygray} G } \;\;,
\eea

In index notation \eqref{eq:defatt3app} resutls in \eqref{eq:new1app}.
\begin{figure*}
\bea \label{eq:new1app}
GSA(K,V,Q)_{q\,a} &=&  \sum_{i=1}^{P}\,\;\; \tfrac{1}{\sqrt{d_q}} \text{softmax} \Big(   \Gamma(Q,K)_{q\,i} \Big)\, {\color{brown}\sum_{j=1}^{P} G_v}_{i j}  \, V_{j\,a}  \;\;\\[0.05cm]  
\text{with} \;\; \bar\Gamma(Q,K)_{q\,i} &=& \sum_{a=1}^{\# heads} \;\; {\color{brown}\sum_{k=1}^{P} G_q}_{q k}\, Q_{k \,a} \,  {\color{brown}\sum_{j=1}^{P} G_k}_{i j}\,[{K}]^T_{\, a \, j}  \;\;,
\\[0.05cm]  
\Gamma(Q,K)_{i\,j}  \, &=& \, \bar\Gamma(Q,K)_{i\,j}  \, + \, \bar\Gamma(Q,K)_{j\,i} \;\;. 
\eea
\end{figure*}
Moreover, \eqref{eq:defatt3app2} in detailed index notation is given by
\bea \nonumber
 \bar\Gamma(Q,K)_{q\,i}\,\;  = \;\;  \,  {\color{blue}\sigma }\Big( \sum_{a=1}^{\# heads}  \,{\color{blue}\sum_{j=1}^{P} G_{qk}}_{q j}\, Q_{ja}\,[\,{K}\,]^T_{\, a \, i} {\color{blue} \, +\, G_{b}}_{qi}  \,\Big) \,.
\eea

\section{Proofs of Invariance}
\label{app:proof}
Permutation matrices $\mathcal{P}$ are orthogonal matrices. An orthogonal matrix is a square matrix whose transpose is equal to its inverse, i.e.\, $\mathcal{P}^T = \mathcal{P}^{-1}$.
A permutation matrix is a square matrix obtained by permuting the rows and columns of an identity matrix. It represents a permutation of the elements in a vector or a rearrangement of the columns and rows of another matrix. Since permuting the rows and columns of an identity matrix results in swapping rows and columns, the transpose of a permutation matrix is equal to its inverse. Therefore, permutation matrices are orthogonal matrices.

The main theoretical claims of this work are summarized in  proposition \ref{thm:sym} , repeated here

\begin{prop}[Symmetry Guarantee]
\label{thm:sym2}
The GSA mechanism~(\eqref{eq:defatt3}) represents a symmetry-preserving module. It may be both invariant and/or equivariant w.r.t. symmetries of the input. The corresponding symmetry is dictated by the various graph selections. To achieve rotation invariance, the subsequent application of \eqref{eq:rot} is necessary. 
   For {\bf invariance}  the  token embedding i.e. the artificial (P-1)$^{th}$ patch is utilized at the output. Due to this mechanism, self-attention~(\eqref{eq:defatt2}) is permutation invariant.
{\bf Equivariance} is achieved for the P-dimensional patch information of the output, i.e.\, not related to the token embedding.

\end{prop}

First of all let us emphasise that we have implicitly empirically tested the validity of this claim in the various RL and supervised experiments in this work
Let us discuss above claims in several steps:

\paragraph{The attention mechanism \eqref{eq:defatt2} is permutation invariant (PI).}
Let $\mathcal{P}$ be a permutation of the input, i.e. $Q,K,V \, \to \, \mathcal{P} Q, \mathcal{P}K,\mathcal{P}V$ as in \eqref{eq:appnew2}.
\begin{figure*}
\bea \label{eq:appnew2}
Att(\mathcal{P}K,\mathcal{P}V,\mathcal{P}Q)_{q\,a} &=&  \sum_{i=1}^{P}\,\;  \text{softmax} \Big(\,\tfrac{1}{\sqrt{d_q}}\, \Sigma(\mathcal{P}Q,\mathcal{P}K)_{q\,i} \, \Big) \, \sum_{j=1}^{P}\mathcal{P}_{ij}V_{j\,a}  \;\;\\[0.05cm]  \nonumber
\text{with} \;\; \Sigma(\mathcal{P}Q,\mathcal{P}K)_{q\,i} &=&  \sum_{a=1}^{\# heads} \;\;\sum_{l=1}^{P}\mathcal{P}_{q l}Q_{l \,a} \Big[\sum_{j=1}^{P}\mathcal{P}_{ij}{K}\Big]^T_{\, a \, j}  \;\;,
\eea
\end{figure*}
with $[\mathcal{P}\,K]^T =K^T \, \mathcal{P}^T \,=  K^T \, \mathcal{P}^{-1}$ and by using that the permutation matrix can be pulled out of the softmax-function as it is not affected by it  results in \eqref{eq:appnew3}.
\begin{figure*}
\bea \label{eq:appnew3}
Att(\mathcal{P}K,\mathcal{P}V,\mathcal{P}Q)_{q\,a} &=&  \sum_{i=1}^{P}\,\;\sum_{l=1}^{P}\mathcal{P}_{q l}\,  \Big(\,\tfrac{1}{\sqrt{d_q}} \text{softmax}\,  \Sigma(Q,K)_{l\,m} \, \Big) \, \sum_{m=1}^{P}\mathcal{P}^{-1}_{mi}\, \sum_{j=1}^{P}\mathcal{P}_{ij}V_{j\,a} \nonumber  \;\;\\[0.05cm]
 &=&  \sum_{j=1}^{P}\,\;\sum_{l=1}^{P}\mathcal{P}_{q l}\,  \text{softmax} \Big(\,\tfrac{1}{\sqrt{d_q}}\, \Sigma(Q,K)_{l\,j} \, \Big) V_{j\,a} \;\;\nonumber  \;\;\\[0.05cm]
  &=& \, \sum_{l=1}^{P}\mathcal{P}_{q l}\, Att(K,V,Q)_{l\,a}
\eea
\end{figure*}
where we have used that $\sum_{i=1}^{P} \mathcal{P}^{-1}_{mi}\,\mathcal{P}_{ij} = \delta_{mj} $, where $\delta$ is the Kronecker delta function, i.e. a formal way of writing the identity matrix. {\bf We have showed that the attention mechanism is permutation equivariant.
Invariance follows form the observation that when a token embedding is added it is not affected by the permutation matrix which only acts on the patches.} Thus it remains invariant as
\beq
Att(\mathcal{P}K,\mathcal{P}V,\mathcal{P}Q)_{q=P+1\,a} = \, Att(K,V,Q)_{q=P+1\,a} \;\;
\eeq

\paragraph{The graph symmetric attention mechanism \eqref{eq:defatt3} represents a symmetry-preserving , i.e. it can be both invariant and equivariant w.r.t. symmetries of the input. The corresponding symmetry is dictated by the various graph selections.} 
Let $\mathcal{P}$ be a permutation of the input ,i.e. $Q,K,V \, \to \, \mathcal{P} Q, \mathcal{P}K,\mathcal{P}V$ then one finds \eqref{eq:newapp4}.
\begin{figure*}
\bea \label{eq:newapp4}
GSA(\mathcal{P}K,\mathcal{P}V,\mathcal{P}Q)_{q\,a} &=&  \sum_{i=1}^{P}\,\;\;  \text{softmax} \Big( \,\tfrac{1}{\sqrt{d_q}}\,  \Gamma(\mathcal{P}Q,\mathcal{P}K)_{q\,i} \Big)\, {\color{brown}\sum_{j=1}^{P} G_v}_{i l} \sum_{l=1}^{P}\mathcal{P}_{lj} \, V_{j\,a}  \;\;\\[0.05cm]  
\text{with} \;\; \Gamma(\mathcal{P}Q,\mathcal{P}K)_{q\,i} &=& \sum_{a=1}^{\# heads} \;\; {\color{brown}\sum_{k=1}^{P} G_q}_{q k}\,\sum_{l=1}^{P}\mathcal{P}_{km}\, Q_{m \,a} \,  {\color{brown}\sum_{j=1}^{P} G_k}_{i j}\,\sum_{l=1}^{P}\mathcal{P}_{jl} \,[{K}]^T_{\, a \, l}  \;\;,
\eea
\end{figure*}
First of all note that since $\mathcal{P} \,\,  {\color{brown} G }  -  {\color{brown} G }  \,\,  \mathcal{P}  \neq 0$ , i.e. they do not commute, thus GSA is not permutation equivariant (invariant).

\begin{definition}[Graph Matrix]
The symmetric graph matrix  $G \in \mathbb{R}^P \times \mathbb{R}^P$ are defined as having a shared weight at entry $G_{ij} = G_{ji}$ if 
\begin{itemize}
\item the distance of the i-vertex to the j-vertex in the square grid of the 2D-image have same length. This leads to a flip and rotation invariant graph matrix.
\item the horizontal distance of the i-vertex to the j-vertex are the same, and the vertical distance is zero. This leads to a horizontal mirror flip graph.
\end{itemize}
\end{definition}
Thus a permutation $\mathcal{P}^s$ does commute with the graph matrix if and only if it maps shared weights of G to each other; then  $\mathcal{P}^s \,\,  {\color{brown} G }  -  {\color{brown} G }  \,\,  \mathcal{P}^s   = 0$.
Then one can rewrite the above expression as in \eqref{eq:appnew5}
\begin{figure*}
\bea \label{eq:appnew5}
GSA(\mathcal{P}^s K,\mathcal{P}^s V,\mathcal{P}^s Q)_{q\,a} &=&  \sum_{i=1}^{P}\,\;\; \text{softmax} \Big(  \,\tfrac{1}{\sqrt{d_q}} \, \Gamma(\mathcal{P}^sQ,\mathcal{P}^sK)_{q\,i} \Big)\, \sum_{l=1}^{P}\mathcal{P}^s_{il} {\color{brown}\sum_{j=1}^{P} G_v}_{lj} \, V_{j\,a} \nonumber  \;\;\\[0.05cm]  
\text{with} \;\; \Gamma(\mathcal{P}^sQ,\mathcal{P}^sK)_{q\,i} &=& \sum_{a=1}^{\# heads} \;\;\sum_{l=1}^{P}\mathcal{P}^s_{qk}\ {\color{brown}\sum_{k=1}^{P} G_q}_{k m}\, Q_{m \,a} \, \sum_{l=1}^{P}\mathcal{P}^s_{ij}  {\color{brown}\sum_{j=1}^{P} G_k}_{ jl}\,\,[{K}]^T_{\, a \, l}  \;\;,
\eea
\end{figure*}
The remaining steps are analog to the one for the conventional attention which is concluded in \eqref{eq:appnew6}.
\begin{figure*}
\bea \label{eq:appnew6}
GSA(\mathcal{P}^s K,\mathcal{P}^s V,\mathcal{P}^s Q)_{i\,a} &=&  \sum_{l=1}^{P}\mathcal{P}^s_{ij} \, GSA( K, V, Q)_{j\,a} \;\;\\[0.05cm]  
GSA(\mathcal{P}^s K,\mathcal{P}^s V,\mathcal{P}^s Q)_{i=P+1\,a} &=&   GSA( K, V, Q)_{i=P+1\,a}
\eea
\end{figure*}
{\bf which states the equivariance of GSA and invariance when a token embedding is used at dimension $P+1$ with respect to the symmetry preserving permutation $\mathcal{P}^s$.} 

Two  remaining points are twofold. First, the derivation above holds in particular  for the symmetrisation (anti-symmetrisation) of $\Gamma$ . Second, to see that graphs matrices with particular choices of shared weights lead to the desired symmetries of the 2D-grid we refer the reader to a visual proof given in figure \ref{fig:gsatt}. The proof of equivariance for the setting with  ${\color{blue} G_{kq}, G_b } $is analog. The case for ${\color{blue} G } $ one notes that a $P\times P$-matrix  $ \sum_{i=1}^P \mathcal{P}^s_{ki} M_{ij} \ast {\color{blue} G }_{ij} =  M_{ij} \ast {\color{blue} G }_{ij}$ if and only if $\mathcal{P}^s$ only permutes shareable weights of the graph matrix, i.e.\, thus $ \mathcal{P}^s$ needs to obey the symmetry properties. The rest of the steps to conclude the proof are analog as above.

\paragraph{\bf To achieve rotation invariance, the subsequent application of \eqref{eq:rot} is necessary. }
Let $  \Gamma_{ij} $  be graph $P\times P$-matrix obeying rotation and flip symmetries of an underlying square grid, and let $\mathcal{T}(i,j) \mapsto k $ be unique mapping for every tuple $(i,j)$ to the vertex $k$; i.e. such that $(i,j,k)$ forms a triangle in the grid. Moreover $\mathcal{T}$ is such that the angles of the resulting triangle only depend on the distance between i-j., and the orientation in clock-wise starting from $i \to j \to k \to i$. The weights $\Theta(ij,jk)$ are shared if the angles in the triangle are equal. 
\bea
  \Gamma^{\text{rot}}(Q,K)_{ij} \;\; =\;\;  \Theta^{(i\to j \to k)}\, \Gamma(Q,K)_{ij} \\[0.2cm] \nonumber
  \, + \,\Theta^{(j\to k \to i)} \,\Gamma(Q,K)_{jk} \, + \, \Theta^{(k\to i \to j)}\, \Gamma(Q,K)_{ki} \;\; ,
  \label{eq:Gammarotf}
 \eea
The notation is such that $ \Theta^{(i\to j \to k)}$ means corresponding to the angle between the edges i-j and j-k.
Then we may rewrite the expression from the main text more concisely as
where $\mathcal{T}(i,j)= k$. Any transformation which maps equal distance edges to each other will leave $\Gamma(Q,K)_{ij}, \Gamma(Q,K)_{jk}, \Gamma(Q,K)_{ki}$ invariant, respectively. However more specifically flip transformation change the meaning of clock-wise and anti-clockwise and thus 
\begin{itemize}
    \item $ \Theta^{(i\to j \to k)} \mapsto \Theta^{(k\to i \to j)}$,
    \item $\Theta^{(j\to k \to i)}\mapsto \Theta^{(j\to k \to i)}$ ,   
    \item $\Theta^{(k\to i \to j)} \mapsto  \Theta^{(i\to j \to k)} $. 
\end{itemize}
Thus \eqref{eq:Gammarotf} does not preserve flip transformation. While rotations leave the $\Theta's$ invariant. This concludes the proof of the proposition.

\section{Hyper-parameters - Main Experiments}
 \label{sec:hype}
 \subsection{\ Minigrid - Lavacrossing}
  We employ our invariant SiTs on top of IMPALA  \cite{espeholt2018impala} implementation based on torchbeast \citep{torchbeast2019}.  For the experiments involving SiTs and ViTs, we do not employ any hyper-parameter tuning compared to the CNN  baseline \cite{jiang2021gtg} - hyperparameters can be found in that reference. 
 We use the stated number of local and global GSA with an embedding dimension of 64 and 8 heads. The attention window of the global GSA is the entire image, i.e $14 \times 14$ (pixels) which corrsponds to $P = 196$ patches ; while locally we choose  patch-size of 5 (pixels), i.e. a attention window of 5x5 (pixels) or  $P = 25$ patches. We use rotation invariant GSA both on the local as well as on  the global level.

 While, MiniGrid environments are partially observable by default  we configure our instances to be fully observable; as well as change the default observation size $9\times 9$ pixel to $14\times 14$ pixel. The latter is done by rendering the environment observation and then down-scaling to the respective size. Moreover, the default action space only allows the agent to turn left, turn right, or move forward, which requires the agent to keep track of its direction while navigating. To make the environment more accessible , we modify the action space, enabling the agent to move in all four candidate directions i.e. to move left, right, forward and backward. 
 
 For Minigrid, we selected a 14x14 global image size as this is the minimum resolution at which the direction of the triangular shaped agent can be discerned when downscaling the RGB rendering of the environment. Smaller resolutions fail to capture this detail, while larger ones are feasible but not necessary. The local neighborhood size in Minigrid (in pixels) is required to be odd, so 3x3 and 5x5 are the smallest viable options.

\textbf{Architecture Details}. For the CNN baseline, we use two convolutions layers two fully connected ones. The ViT and SiT and architectures both employ a skip connection $x\ \, \to \,x \, + \, \text{Att(x)}$ and $x\ \, \to \, x \, + \, \text{GSA(x)}$, respectively.  As in  \citep{beyer2022better}, we modify the Vi architecture of \cite{vit} by not using a multi-layer perceptron (MLP) after each attention layer. Our goal is to encounter the most simple functional setting incorporating the attention mechanism.  We use one embedding fully-connected layer and use a patch-embedding. We employ our invariant SiTs on top of IMPALA  \cite{espeholt2018impala} implementation based on torchbeast\citep{torchbeast2019}.  For the experiments involving SiTs and ViTs, we do not employ any hyper-parameter tuning compared to the CNN  baseline \cite{jiang2021gtg}. Qualitatively, even when trying to tune ViTs by running a hyper-parameter sweep, we could not improve their performance by more than a factor of two.

 \subsection{ Procgen experiments}

We train with PPO (DrAC) + Crop augmentation for SiT (Sit$^\ast$, Set, Siet) and compare to the CNN (ResNet with model size $79.4k$   comparable to the SiT - $ 65.7k$;  ViT with 4-layers - $ 216k$ ). For parameter details see table~(2).

We do not alter the ResNet architecture of \cite{raileanu2020automatic} but chose the same hidden-size of 64 as for the SiT as well reduce the number of channels to [4,8,16]. We train over 25M steps. Following \citet{agarwal2021deep}, we report the min-max normalized score that shows how far we are from maximum achievable performance on each environment. All scores are computed by averaging over both the 4 seeds and over the  23M-25M test-steps. We also report UCB-DrAC results with Impala-CNN($\times4$) ResNet with 620k parameters, taken from \cite{raileanu2020automatic}.

 \subsection{ DM control}
\label{app:dmc}
 
We train  SieT - Without any hyper-parameter and backbone changes compared to our Procgen setup - wit SAC \cite{SAC} for 500k steps. No data-augmentation is used. SieT has comparable performance to the ViT baseline with $>$ 1M weights~\citep{hansen2021softda}.  For parameter details see table~(2).

\begin{table*}[t!]
    \centering
    \resizebox{2\columnwidth}{!}{
    \begin{tabular}{c|c|c|c|c}
        \toprule
        \textbf{Suite / Algorithm} & \textbf{Model} & \textbf{Layers} & \textbf{Channels} & \textbf{Hidden dim.} \\
        \midrule
        \multirow{9}{*}{\textbf{Procgen} \cite{raileanu2020automatic}}
        & \textit{CNN} & 3 ResNet \cite{raileanu2020automatic} & [4,8,16] & 64 \\
        & \textit{ViT} & 4 Attn. Layers \cite{hansen2021softda} & 64 & 64 \\
        & \textit{E2CNN} & 4 E2CNN Layers \cite{wang2022mathrmsoequivariant} & [2,4,8,8] & 64 \\
        & \textit{E2CNN$^\prime$} & 4 E2CNN Layers \cite{wang2022mathrmsoequivariant} & [2,4,8,16] & 64 \\[0.1cm]
        & SiT & 2 local GSA, 2 global GSA & 64 & \\
        & SiT$^\ast$ & 2 local GSA, 2 global GSA & 64 & 64 \\
        & SeT & 2 local GSA, 2 global GSA & 64 & 64 \\
        & SieT & 2 local GSA, 2 global GSA & 64 & 64 \\[0.1cm]
        & \textcolor{gray}{UCB-CNN} & 3 ResNet \cite{raileanu2020automatic} & [16,32,32] & 256 \\

        \midrule
        \midrule
        \textbf{DM-control} \cite{hansen2021softda} & SieT & 2 local GSA, 2 global GSA & 64 & 64 \\

        \midrule
        \midrule
        \multirow{2}{*}{\textbf{Atari 100k} \cite{hessel2017rainbow}}
        & CNN & data-efficient \cite{hessel2017rainbow} & [32,64] (approx. 880k) & 256 \\[0.1cm]
        & SieT & 1 local GSA, 2 global GSA & 128 (approx. 309k) & 256 \\
        \bottomrule
    \end{tabular}
    }
    \caption{Architecture and hyper-parameter choices for Procgen and DM-control. UCB-CNN is taken from \cite{raileanu2020automatic}. All SiT variants contain one initial conv. layer with shared graph weights and a subsequent max. pooling layer to reduce the dimensionality of the problem. For the Atari experiments, we use the unchanged hyper-parameter setting of Rainbow \cite{hessel2017rainbow}.}
    \label{tab:hyp2}
\end{table*}

\subsection{Proof of sample-efficiency: Atari 100k}
\label{app:atari}

The Atari 100k benchmark \cite{kaiser2020}, comprising 26 Atari games \cite{bellemare13arcade}, spans various mechanics and evaluates a broad spectrum of agent capabilities. In this benchmark, agents are restricted to 100k actions per environment, approximating 2 hours of human gameplay. For context, typical unconstrained Atari agents undergo training for 50 million steps, signifying a 500-fold increase in experience.

Current standards in Atari 100k for search-based methods include MuZero \cite{schrittwieser2020} and EfficientZero \cite{ye2021}, and recently a transformer based world-model approach \cite{iris2023}. Image-based SiTs may compliment those methods. In particular the latter may benefit from GSA, see our discussion in section \ref{secapp:1D}. We just provide a proof-of concept for using shift symmetric GSA instead of positional embedding in the world-model transformer of IRIS  \cite{iris2023}. 

\begin{figure}[t]
    \centering
    \includegraphics[width=0.44\textwidth]{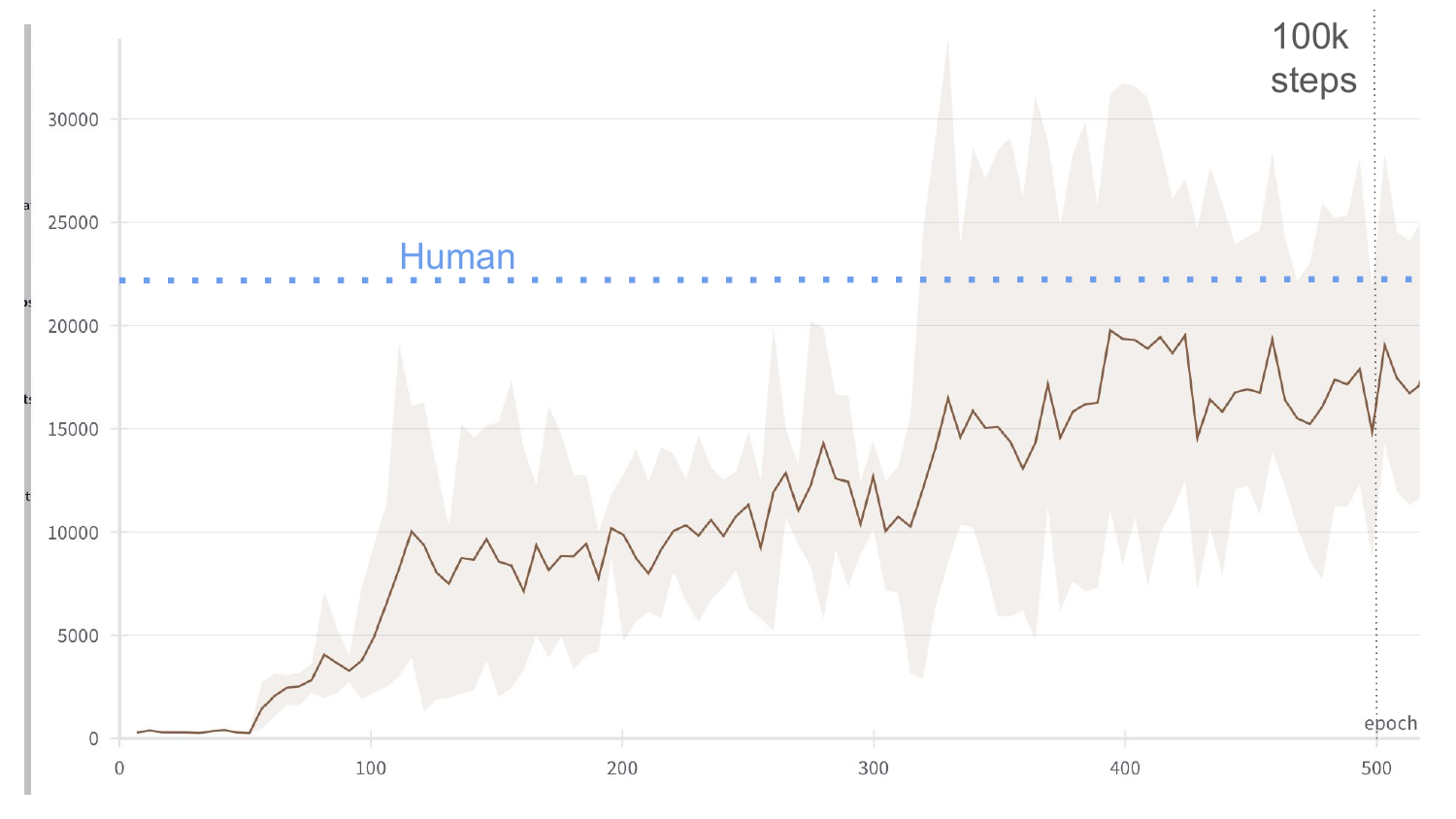}
    \vspace{-0.2cm}
    \caption{Results on KungFuMaster with 1D GSA in \cite{iris2023} averaged over a hyper-parameter search for varying world-model training steps in $[50,200]$ (with default of 200). 500 Epochs correspond to 100k exploration steps.}
    \vspace{-0.1cm}
    \label{fig:IRIS}
\end{figure}

See Figure~\ref{fig:IRIS} for the results on KungFuMaster averaged over a hyper-parameter search for varying world-model training steps for 7 run for values in $[50,200]$ (with default of 200). This illustrated the hyper-parameter insensitivity of GSA and suggests improved sample-efficiency over positional embedding. 

\begin{figure}[t]
    \centering
      \vspace{-0.8cm}
    \includegraphics[width=0.5\textwidth]{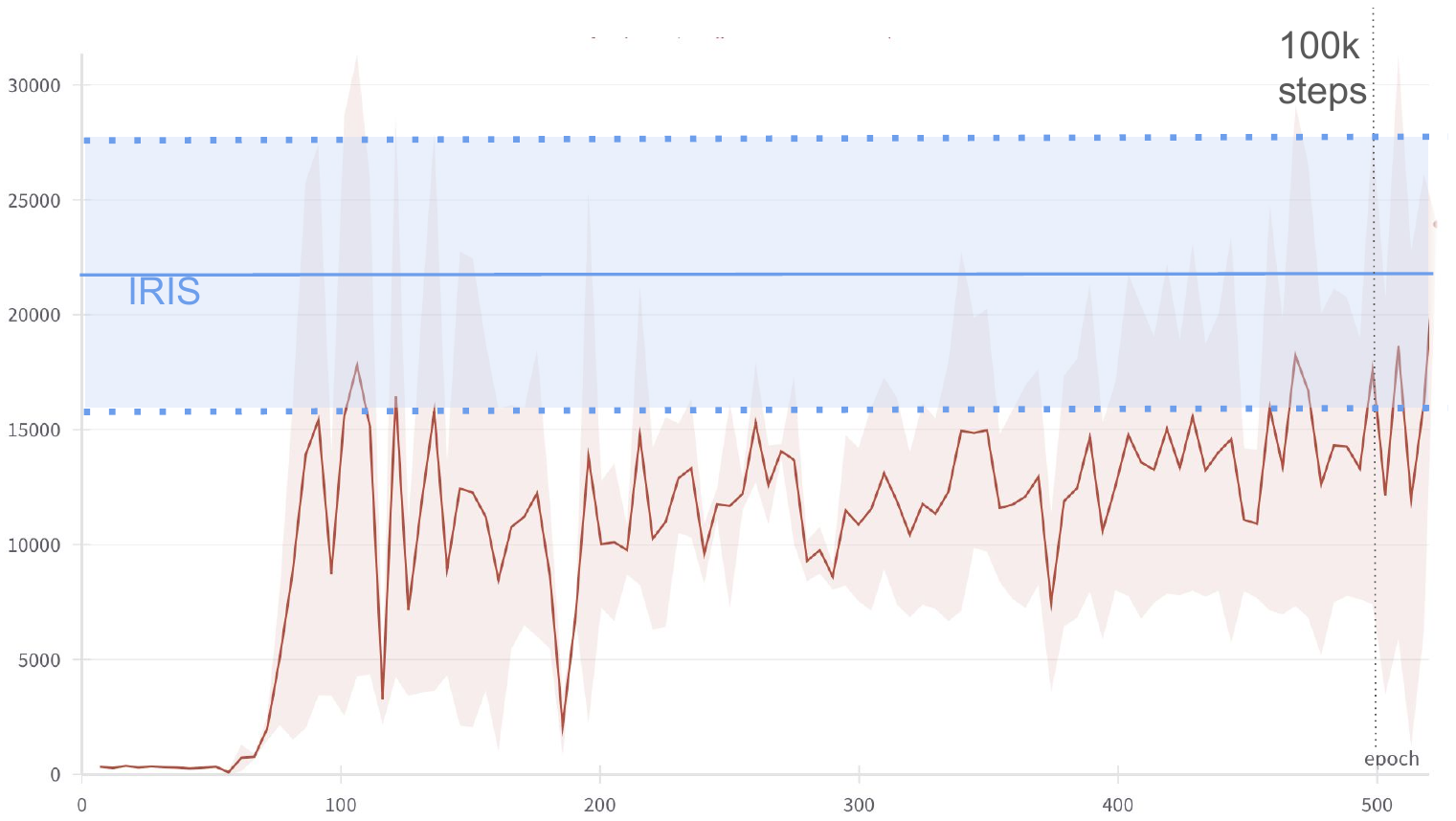}
    \vspace{-1.4cm}
    \caption{Results on KungFuMaster with SieT in actor instead of CNN in  \cite{iris2023}, averaged over 3 seeds. The IRIS performance after 100k steps is marked by a blue horizontal bar.}
    \vspace{-0.1cm}
    \label{fig:IRIS2}
\end{figure}

\begin{figure}[t]
    \centering
      \vspace{-0.8cm}
    \includegraphics[width=0.5\textwidth]{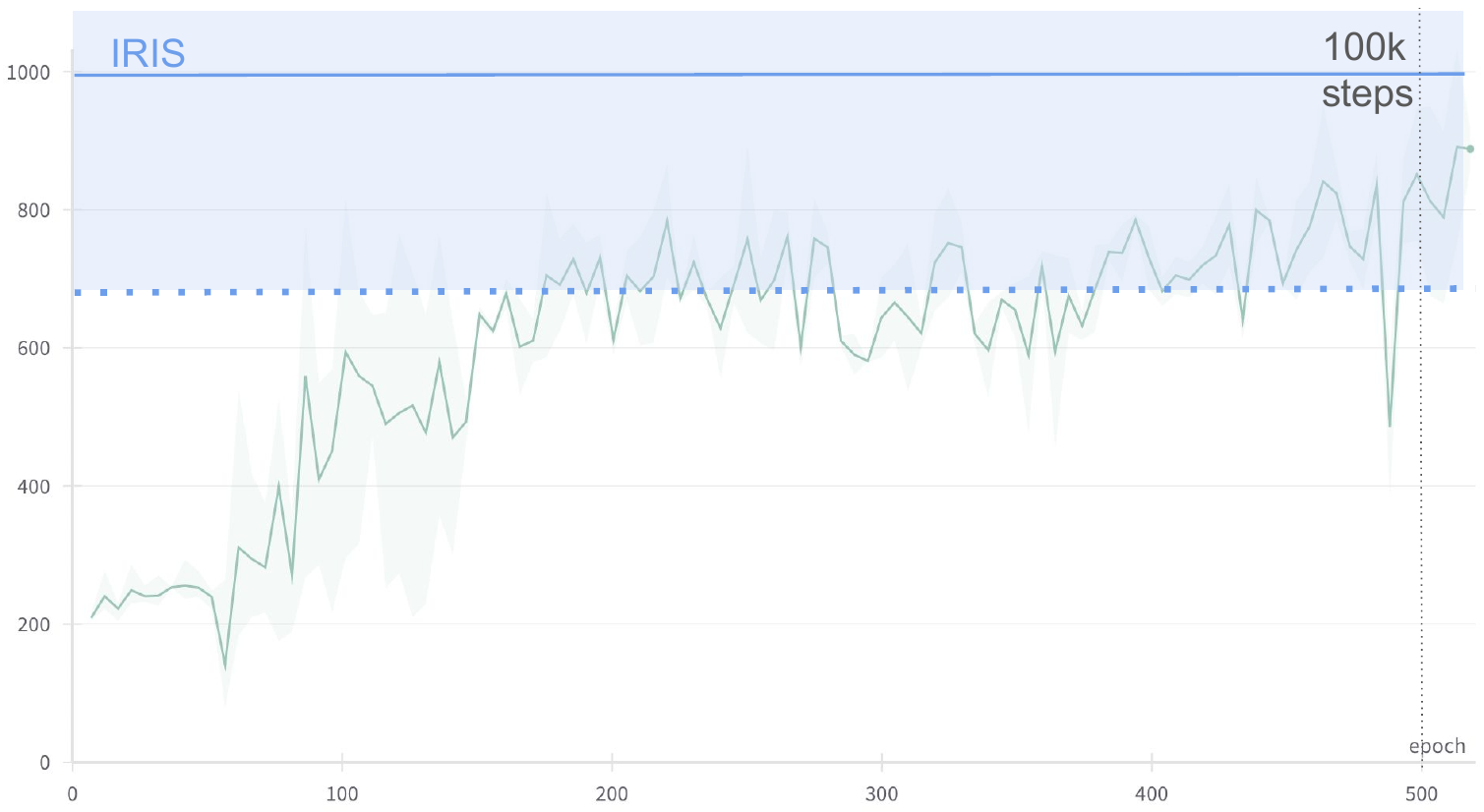}
    \vspace{-1.4cm}
    \caption{Results on MsPacMan with SieT in actor instead of CNN in  \cite{iris2023}, averaged over 3 seeds. The IRIS performance after 100k steps is marked by a blue horizontal bar. }
    \vspace{-0.1cm}
    \label{fig:IRIS3}
\end{figure}

{\bf Sample-efficiency:} See Figure~\ref{fig:IRIS2} and Figure~\ref{fig:IRIS3} for results on KungFuMaster and MsPacMan, respectively. Here we have replaced the CNN in the actor by a smaller SieT model with 1 local and 2 global layers. To speed up training we have reduced the number of local features to 32 ( 64 global features). One infers that although the final training performance is lower than the one obtained by the baseline (IRIS + CNN) that IRIS + SieT is sample efficient, i.e. it  reaches it peak  quite early in training within 200 epochs which corresponds to about 40k exploration steps. Note that we have only tested one SieT model and performed no hyper-parameter search. The goal here is not to improve on Iris which had optimized h-parameters for their CNN actor model but to show that SieT admits the capability to learn from a small sample size; which is inferred from   Figure~\ref{fig:IRIS2} and Figure~\ref{fig:IRIS3}.

\begin{figure}[t]
    \includegraphics[width=0.5\textwidth]{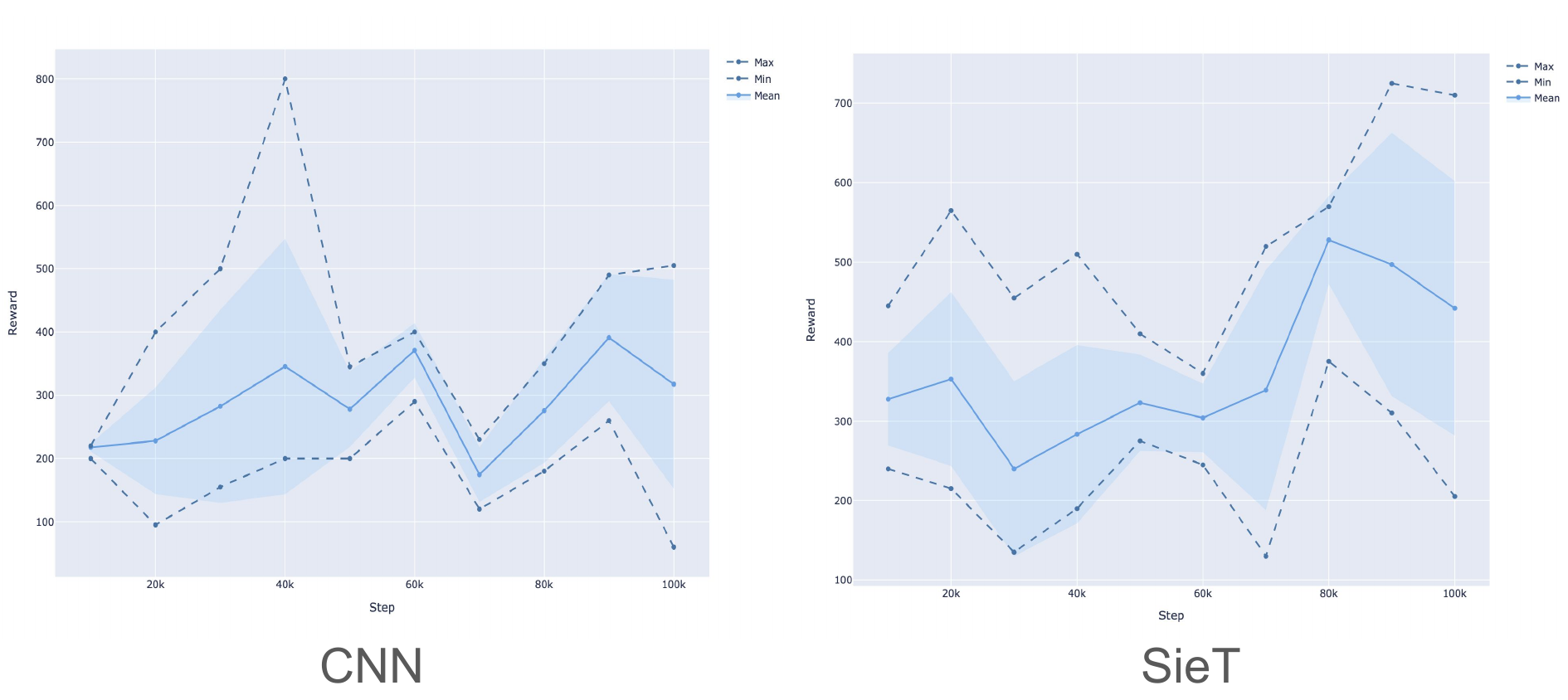}
    \vspace{-0.2cm}
    \caption{Example evaluation curves of Rainbow during training of SpaceInvaders seed 124 on top of rainbow - data efficient -.}
    \label{fig:table_space}
    \vspace{-0.5cm}
\end{figure}

Our goal here is not to show a comparison to state-of-the-art on Atari 100k in terms of sample-efficiency but that SiTs may be trained with ease on-top of standard algorithms.
Thus, for our Atari 100k experiments  we {\bf use the unchanged hyper-parameter setting of Rainbow \cite{hessel2017rainbow}}, both for the baseline model - data-efficient CNN  - as well as the algorithm for both CNN and our SieT. The Siet model is the same as in the Procgen experiments but with one less local GSA layer and increased embedding dimension, see Table~2. {\bf The baseline CNN has approximately 3x more weights than SieT. See Figure~\ref{fig:table_space} for example training curves.}

\begin{table}[t]
\centering
\resizebox{0.6\columnwidth}{!}{
\begin{tabular}{ l |  c || c  }
\toprule
\multicolumn{3}{c}{Rainbow - data-efficient}\\
\midrule
\midrule
\textbf{Atari Game } & \textbf{CNN}  & \textbf{SieT}  \\
\midrule
SpaceInvaders & 344 & 366.3  \\
BreakOut & 4.3 &4.97  \\
Pong  & -19.07 & -20.04 \\
\bottomrule 
\end{tabular}
}
\caption{\textbf{CNN vs. SiTs on Atari 100k benchmark environments: SpaceInvaders, BreakOut, Pong}. 
We train with Rainbow \cite{hessel2017rainbow} and compare to the data-efficient CNN ($\approx 880k$ weights ) about $3 \times$ larger than our SieT - $ \approx 309k$. The number model parameters vary for different environments, however the factor in between CNN baseline and SieT is consistently $ \approx 3 \times$. We present absolute scores, averaged over evaluation after 80k,90k,100k train-steps and 3 seeds, respectively.
}
    \label{tab:atari}
\end{table}

 \subsection{ Technical improvements }
Moreover,  another minor modification to the architecture lead to a compute speed-up, which is not to use a token embedding in the local attention layer but rather use a depth wise graph-like convolution with kernel-size and stride equal to the patch-size as a last layer. This is common practice for ViTs i.e. by using normal convolutions that way. It is easy to see that our choice also preserves the symmetries of the graph-matrix. So, concludingly using a token embedding is not the only architectural choice which leads to a preservation of symmetries in SiTs.

First, we  establish a connection between graph matrices and depth-wise convolutions with graph-weights as kernels. The latter implementation is significantly more memory efficient and faster.
Secondly, in order to accommodate for an extend local attention window we do use a graph matrix which connects pixels over  lager distances while keeping the actual attention-mechanism focused on a smaller patch.

Thirdly, rather trivially one may scale down the original image size from $64 \times 64$ pixels to $32 \times 32$. This can be done by simple scaling the image, or by using one initial depth-wise convolutional layer with  graph-like weights to preserve symmetry plus a subsequent  Max-Pooling operation, or by simply using every second pixel of the input-image. 

Given the dominance of ViTs in Vision and Transformers in NLP, it's plausible that improvements in Transformer technology will similarly revolutionize vision-based RL, with ViTs becoming predominant. Recent technical advancements, such as efficient Transformers \cite{dao2022flashattention}, which offer up to a 10x performance boost may lead the way to a brad adaption of SiTs in RL. As the latter ensures that our  sample efficient symmetry-invariant  vision transformer becomes rather light-weight.  

\paragraph{\bf CIFAR10 - supervised ablation study:} For the baseline we use a ViT \cite{vit} , embedding dimension of 512, 4 layers, 16 heads, and one local attention layer with the same settings. The SiT has a the graph matrix $ {\color{mygray} G}$ added.

\section{Additional Experiments and Ablations}

\subsection{Ablation: Local and Global Symmetries in Attention for Image Understanding}
\begin{figure}[t]
 \centering
   \includegraphics[width=0.5\textwidth]{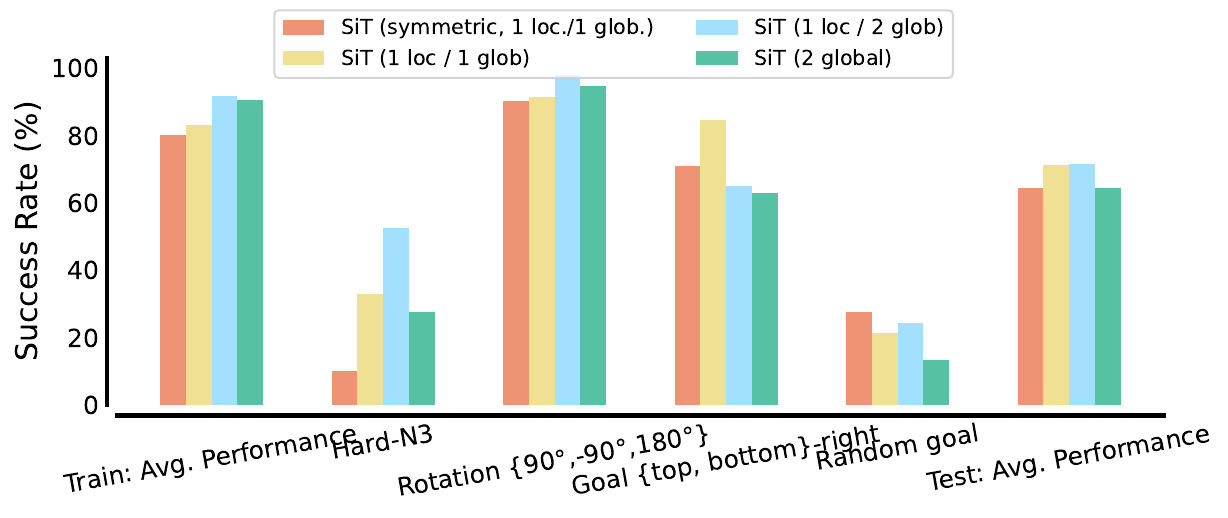}
    \vspace{-0.2cm}
    \caption{Ablations study SiT for different number of local and global layers as well as using the symmetric part of SiT$^\ast$, defined in the appendix.}
    \label{fig:table_ablation}
\end{figure}

In this section we present an ablation study  which addresses the impact of local attention fields in SiTs, see results in Figure~\ref{fig:table_ablation}. For the performance comparison of the attention mechanism with both symmetrisation and anti-symmetrisation (Equation~\ref{eq:defatt4}) in SiTs,  we restrict ourselves to provide qualitative results. 
Performing several experiments with two global layers, of either symmetrisation and anti-symmetrisation or both present, we conclude that indeed the later admits the best relative performance as well as generalisation to the hard task, of all SiT setup with only global layers. However, due to the second softmax in the attention it is more memory consuming and thus applying it in combination with a the local layers is not feasible. Memory efficient attention mechanism \cite{dao2022flashattention} have been developed recently thus our  better architecture may become technically available in the future.

\begin{figure}[t]
    \centering
     \includegraphics[width=0.5\textwidth]{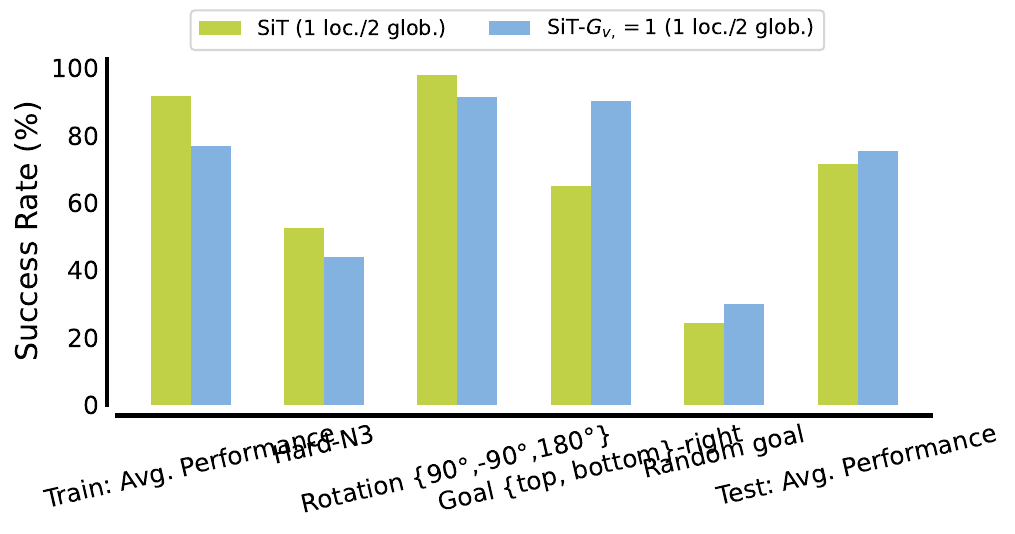}
    \caption{Ablations for using the symmetric part of SiT$^\ast$, defined in the appendix. We ablate by $G_{v,}$ by setting ${\color{brown} G_{v,}} = 1 $ with one local and two global  symmetric attention modules \eqref{eq:defatt4} and rotation invariance.}
    \label{fig:ablate_gv}
\end{figure}

Finally, we compare the impact of the employment of graph matrices in the values of the attention mechanism. We set ${\color{brown} G_{v,}} = 1 $ and use one local and two global symmetric attention modules~(Equation~\ref{eq:defatt4}) with rotation invariance, see Figure~\ref{fig:ablate_gv}. The graph matrix  ${\color{brown} G_{v,}} $ leads to a decrease in generalisation performance on goal change tasks and an increase in performance on the difficult environment involving many lava-crossings.   Conceptually,  the graph-matrix  ${\color{brown} G_{v,}} $ admits some similarities to convolutions, which are known to preform better on limited sample size~\citep{hassani2022escaping}. Thus it may be easier for this architecture to identify two-dimensional information while setting it to the identity restores some degree of permutation invariance.

\subsection{Minigrid - Lavacrossing}
In this section we briefly provide the details with error bars to the experiment provided in the main text. The tables \ref{tab:table_grd}, \ref{tab:table_ablation}  contain the mean rewards  averaged over 200 test episodes after 20M time-steps.

\label{app:abl}
\begin{table*}[t!]
    \centering
    \caption{Details main results for the experiment on the Lavacrossing Minigrid environment.}
    \resizebox{2\columnwidth}{!}{
    \begin{tabular}{l|l|cccccc}
        \toprule
        \textbf{Domain Train} & \textbf{Test Task} & \textbf{CNN} & \textbf{ViT} & \textbf{PI-ViT} & \textbf{SiT} & \textbf{SiT-1} \\
        \midrule
        \multirow{7}{*}{\shortstack{\textbf{Lavacrossing N1 \& N2}\\\textbf{medium \& easy}}} 
        & \textit{train: easy-N1} & $0.96 \pm 0.01$ & $0.23 \pm 0.41$ & $0.01 \pm 0.0$ & $0.94 \pm 13.0$ & $0.90 \pm 0.23$ \\
        & \textit{train: medium-N2} & $0.69 \pm 0.43$ & $0.25 \pm 0.42$ & $0.01 \pm 0.0$ & $0.84 \pm 0.32$ & $0.70 \pm 0.42$ \\
        & hard-N3 & $0.54 \pm 0.47$ & $0.19 \pm 0.38$ & $0.01 \pm 0.0$ & $0.46 \pm 0.48$ & $0.41 \pm 0.47$ \\
        & rotations average & $0.00 \pm 0.00$ & $0.15 \pm 0.34$ & $0.01 \pm 0.0$ & $0.91 \pm 0.21$ & $0.91 \pm 0.22$ \\
        & goal top-right & $0.08 \pm 0.27$ & $0.21 \pm 0.40$ & $0.01 \pm 0.0$ & $0.71 \pm 0.44$ & $0.90 \pm 0.24$ \\
        & goal bottom-left & $0.08 \pm 0.25$ & $0.20 \pm 0.40$ & $0.01 \pm 0.0$ & $0.71 \pm 0.44$ & $0.90 \pm 0.24$ \\
        & random goal & $0.07 \pm 0.25$ & $0.15 \pm 0.35$ & $0.01 \pm 0.0$ & $0.24 \pm 42.0$ & $0.21 \pm 0.40$ \\
        \bottomrule
    \end{tabular}
    }
    \label{tab:table_grd}
\end{table*}

\begin{table*}[t!]
    \centering
    \caption{Ablation study for SiT with different numbers of local and global layers as well as using the symmetric part of SiT$^\ast$, defined in the appendix.}
    \resizebox{2\columnwidth}{!}{
    \begin{tabular}{l|l||cc|cccc}
        \toprule
        \textbf{Train} & \textbf{Test Task} & \multicolumn{2}{c|}{\textbf{ViT}} & \multicolumn{4}{c}{\textbf{SiT}} \\
        \midrule
        \multicolumn{2}{c|}{\text{}} & \text{2 global} & \text{1 loc./2 glob.} & \text{symmetric 1/1} & \text{1 loc./1 glob.} & \text{1 loc./2 glob.} & \text{2 global} \\
        \midrule
        \multirow{7}{*}{\textbf{Lavacr. N1/N2}} 
        & \textit{train: easy-N1} & \text{0.08 $\pm$ 0.27} & \text{0.23 $\pm$ 0.41} & \text{0.86 $\pm$ 0.29} & \text{0.90 $\pm$ 0.23} & \textbf{\text{0.93 $\pm$ 18.0}} & \text{0.87 $\pm$ 0.28} \\
        & \textit{train: medium-N2} & \text{0.12 $\pm$ 0.31} & \text{0.25 $\pm$ 0.42} & \cellcolor{lightyellow}\text{0.88 $\pm$ 0.26} & \text{0.70 $\pm$ 0.42} & \cellcolor{lightyellow}\text{0.84 $\pm$ 0.32} & \cellcolor{lightyellow}\text{0.59 $\pm$ 0.46} \\
        & \text{hard-N3} & \text{0.10 $\pm$ 0.29} & \text{0.19 $\pm$ 0.38} & \cellcolor{lightyellow}\text{0.45 $\pm$ 0.43} & \text{0.41 $\pm$ 0.47} & \cellcolor{lightyellow}\text{0.46 $\pm$ 0.48} & \cellcolor{lightyellow}\text{0.03 $\pm$ 0.16} \\
        & \text{rotations avg.} & \text{0.03 $\pm$ 0.16} & \text{0.15 $\pm$ 0.34} & \text{0.88 $\pm$ 0.27} & \textbf{\text{0.91 $\pm$ 0.22}} & \textbf{\text{0.95 $\pm$ 0.10}} & \text{0.91 $\pm$ 0.21} \\
        & \text{goal top-right} & \text{0.03 $\pm$ 0.15} & \text{0.21 $\pm$ 0.40} & \text{0.81 $\pm$ 0.35} & \textbf{\text{0.90 $\pm$ 0.24}} & \textbf{\text{0.71 $\pm$ 0.44}} & \text{0.60 $\pm$ 0.48} \\
        & \text{goal bottom-left} & \text{0.02 $\pm$ 0.13} & \text{0.20 $\pm$ 0.40} & \text{0.81 $\pm$ 0.35} & \textbf{\text{0.90 $\pm$ 0.24}} & \textbf{\text{0.71 $\pm$ 0.44}} & \text{0.60 $\pm$ 0.48} \\
        & \text{random goal} & \text{0.08 $\pm$ 0.27} & \text{0.15 $\pm$ 0.35} & \cellcolor{lightyellow}\textbf{\text{0.26 $\pm$ 0.43}} & \cellcolor{lightyellow}\textbf{\text{0.21 $\pm$ 0.40}} & \cellcolor{lightyellow}\textbf{\text{0.28 $\pm$ 0.44}} & \cellcolor{lightyellow}\text{0.12 $\pm$ 0.32} \\
        \bottomrule
    \end{tabular}
    }
    \label{tab:table_ablation}
\end{table*}

\section{Additional Evaluations \& Ablations}
\label{app:ps-sit}

\subsection{ Patch-size Ablation on CIFAR 10 }
In this section, we perform an ablation study on the sensitivity of  Sit (Siet) to the  patch-size hyper-parameter. In contrast to the evaluation of Figure~(\ref{fig:se_cifar})  here we employ a much smaller and faster model with 4 local and 4 global GSA layers and have reduced the feature dimensions to 64 and 256, respectively. 

We perform an ablation study of varying patch-size of SieT model on CIFAR10 with resolution of 32x32 and  128x128 pixels, see Tables~(\ref{fig:c32}) and (\ref{fig:c128}), respectively. Additionally, we vary the dimension of the graph matrix in the local GSA for fixed patch-size in selected cases. We conclude that SieTs can be applied without loss of  accuracy performance to higher-resolution tasks and moreover are relatively insensitive to changes in the patch-size \cite{beyer2023flexivit}. 



\begin{table*}[ht]
\centering
\caption{CIFAR 10 - 32x32 Results. Ablation study of varying patch-size of SieT model on CIFAR10 with resolution of 32x32 pixels. Additionally, we vary the dimension of the graph matrix in the local GSA, see x-axis. Gray bars indicate non-available cases, i.e. the combination of graph-matrix size with patch-size has not been conducted. }
\vspace{0.2cm}
  \resizebox{2\columnwidth}{!}{
\begin{tabular}{c|c|c|c|c|c|c}
 \toprule
  \textbf{GSA Layers} & \textbf{Patch-Size Global} & \textbf{Patch-Size Local} & \textbf{Graph Matrix Size} & \textbf{Batch-Size} & \textbf{Test Accuracy 25 Epochs} & \textbf{Test Accuracy 200 Epochs} \\ 
  \midrule
4loc-4glob & 8x8 & 4x4 & 8x8 & 64 & 66\% & - \\ 
4loc-4glob & 4x4 & 8x8 & 16x16 & 64 & 62\% & - \\ 
4loc-4glob & 16x16 & 2x2 & 4x4 & 64 & 66\% & - \\ 
4loc-4glob & 4x4 & 8x8 & 8x8 & 64 & 58\% & 75\% \\ 
4loc-4glob & 16x16 & 2x2 & 8x8 & 64 & 67\% & 81\% \\ 
4loc-4glob & 8x8 & 4x4 & 6x6 & 64 & 65\% & - \\ 
4loc-4glob & 8x8 & 4x4 & 8x8 & 64 & 66\% & - \\ 
4loc-4glob & 8x8 & 4x4 & 12x12 & 64 & 66\% & - \\ 
 \bottomrule
\end{tabular}
}
\label{fig:c32}
\end{table*}

\begin{table*}[ht]
\centering
\caption{CIFAR 10 - 128x128 Results. Ablation study of varying patch-size of SieT model on CIFAR10 with resolution of 128x128 pixels. The images are scaled up using AI tools \href{https://www.kaggle.com/datasets/joaopauloschuler/cifar10-128x128-resized-via-cai-super-resolution}{CIFAR-10 128x128}.
 Additionally, we vary the dimension of the graph matrix in the local GSA, see x-axis. Gray bars indicate non-available cases, i.e. the combination of graph-matrix size with patch-size has not been conducted.}
 \vspace{0.2cm}
 \resizebox{2\columnwidth}{!}{
\begin{tabular}{c|c|c|c|c|c|c}
\toprule
 \textbf{GSA Layers} & \textbf{Patch-Size Global} & \textbf{Patch-Size Local} & \textbf{Graph Matrix Size} & \textbf{Batch-Size} & \textbf{Test Accuracy 25 Epochs} & \textbf{Test Accuracy 15 Epochs} \\ 
 \midrule
4loc-4glob & 8x8 & 16x16 & 16x16 & 48 & 66\% & 63\% \\ 
4loc-4glob & 8x8 & 16x16 & 32x32 & 48 & 69\% & 64\% \\ 
4loc-4glob & 8x8 & 16x16 & 64x64 & 48 & 69\% & 64\% \\ 
4loc-4glob & 8x8 & 16x16 & 32x32 & 16 & 70\% & 66\% \\ 
4loc-4glob & 4x4 & 32x32 & 64x64 & 8 & 69\% & 66\% \\ 
 \bottomrule
\end{tabular}
}
  \label{fig:c128}
\end{table*}


\end{document}